\documentclass[conference]{IEEEtran}
\IEEEoverridecommandlockouts

\newif\ifacmartsty
\acmartstyfalse
\newif\ifcameraready
\camerareadytrue

\ifacmartsty
\else
\usepackage[
  pass,
]{geometry}
\fi

\usepackage[T1]{fontenc}
\usepackage[utf8]{inputenc}
\usepackage{upquote}

\usepackage{microtype}
\UseMicrotypeSet[protrusion]{basicmath} 

\usepackage{hyperref}
\hypersetup{
  colorlinks   = true, 
  urlcolor     = blue, 
  linkcolor    = red, 
  citecolor   = blue 
}
\usepackage{graphicx,grffile}
\usepackage{times}
\usepackage{multirow}
\usepackage{xspace}
\usepackage{tabularx}
\usepackage{ragged2e}
\usepackage{booktabs}
\usepackage{paralist}
\usepackage[american]{babel}
\usepackage[shortlabels]{enumitem}
\usepackage{courier,color,wrapfig}
\usepackage{xspace}
\usepackage{balance}
\usepackage{enumitem}
\usepackage{epstopdf}
\usepackage{multirow}
\usepackage{booktabs}
\usepackage{url}
\usepackage{pifont}
\usepackage{subfigure}
\usepackage{tikz}
\usepackage{mathtools}
\usepackage[normalem]{ulem}
\usepackage{cite}

\usepackage[linesnumbered,lined,plain]{algorithm2e}
\usepackage{etoolbox}
\usepackage[textsize=small]{todonotes}
\usepackage[compact]{titlesec}
\usepackage{caption}
\SetAlgoCaptionSeparator{.}
\usepackage[export]{adjustbox}

\usepackage{gensymb} 
\usepackage[capitalize]{cleveref}
\crefformat{section}{§#2#1#3}
\crefname{enumi}{Step}{Steps}
\crefname{algorithm}{Alg.}{Algs.}

\makeatletter
\def\maxwidth{\ifdim\Gin@nat@width>\linewidth\linewidth\else\Gin@nat@width\fi}
\def\maxheight{\ifdim\Gin@nat@height>\textheight\textheight\else\Gin@nat@height\fi}
\makeatother
\setkeys{Gin}{width=\maxwidth,height=\maxheight,keepaspectratio}
\makeatletter
\g@addto@macro{\UrlBreaks}{\UrlOrds}
\makeatother

\ifacmartsty
\makeatletter
\let\origsection\section
\let\origsubsection\subsection

\renewcommand\section{\@ifstar{\starsection}{\nostarsection}}
\renewcommand\subsection{\@ifstar{\starsubsection}{\nostarsubsection}}

\newcommand\sectionprelude{\vspace{1.5ex}}
\newcommand\sectionpostlude{\vspace{1.5ex}}
\newcommand\subsectionprelude{\vspace{0.5ex}}
\newcommand\subsectionpostlude{\vspace{0.5ex}}

\newcommand\nostarsection[1]{\sectionprelude\origsection{#1}\sectionpostlude}
\newcommand\starsection[1]{\sectionprelude\origsection*{#1}\sectionpostlude}

\newcommand\nostarsubsection[1]{\subsectionprelude\origsubsection{#1}\subsectionpostlude}
\newcommand\starsubsection[1]{\subsectionprelude\origsubsection*{#1}\subsectionpostlude}

\makeatother
\fi

\newcommand\paraspace{\vspace*{0.2ex}}
\providecommand\parab[1]{\paraspace\noindent\textbf{#1}}
\providecommand\parae[1]{\paraspace\textbf{\textit{#1}}}

\apptocmd\normalsize{%
\abovedisplayskip=5pt
\abovedisplayshortskip=5pt
\belowdisplayskip=5pt
\belowdisplayshortskip=5pt
}{}{}

\ifacmartsty
\renewcommand\footnotetextcopyrightpermission[1]{} 
\setcopyright{none}
\acmDOI{}
\acmISBN{}
\acmConference[Submitted for review]{}
\acmYear{2018}
\copyrightyear{}
\acmPrice{}
\fi

\newcommand{\sysname}{CIP\xspace}

\newcommand{\etc}{\emph{etc.}\xspace}
\newcommand{\ie}{\emph{i.e.,}\xspace}
\newcommand{\eg}{\emph{e.g.,}\xspace}

\newlist{myitemize}{itemize}{1}
\setlist[myitemize]{nosep,leftmargin=*}
\newlist{myenumerate}{enumerate}{1}
\setlist[myenumerate]{nosep,leftmargin=*,label=\textbf{\arabic*}.}
\newlist{mydescription}{description}{1}
\setlist[mydescription]{nosep,leftmargin=*}

\newcommand{\secref}[1]{\S\ref{#1}}
\newcommand{\figref}[1]{Fig.~\ref{#1}}

\newcommand{\algoref}[1]{Algorithm~\ref{#1}}

\newcommand{\fawad}[1]{\todo[author=Fawad,color=orange,inline]{#1}}
\newcommand{\ramesh}[1]{\todo[author=Ramesh,color=blue!25,inline]{#1}}
\newcommand{\weiwu}[1]{\todo[author=Weiwu,color=yellow,inline]{#1}}
\newcommand{\christina}[1]{\todo[author=Christina,color=pink,inline]{#1}}

\newcommand{\footnoteref}[1]{\textsuperscript{\ref{#1}}}

\usepackage{authblk}

\begin{document}

\title{Cooperative Infrastructure Perception}

\author[1]{Fawad Ahmad*\thanks{* Equal contribution to this work.}}
\author[2]{Christina Suyong Shin*}
\author[2]{Weiwu Pang*}
\author[2]{Branden Leong}
\author[3]{Pradipta Ghosh}
\author[2]{Ramesh Govindan}
\affil[1]{Rochester Institute of Technology, Rochester, NY, USA}
\affil[2]{University of Southern California, Los Angeles, CA, USA}
\affil[3]{Meta, Menlo Park, CA, USA}
\affil[ ]{fawad@cs.rit.edu, \{cshin956, weiwupan, branden, ramesh\}@usc.edu, iampradipta@meta.com}


\maketitle


\begin{abstract}

Recent works have considered two qualitatively different approaches to overcome line-of-sight limitations of 3D sensors used for perception: cooperative perception and infrastructure-augmented perception.
In this paper, motivated by increasing deployments of infrastructure LiDARs, we explore a third approach~--~\textit{cooperative infrastructure perception}.
This approach generates perception outputs by fusing outputs of multiple infrastructure sensors, but, to be useful, must do so quickly and accurately.
We describe the design, implementation and evaluation of Cooperative Infrastructure Perception (\sysname), which uses a combination of novel algorithms and systems optimizations.
It produces perception outputs within 100~ms using modest computing resources and with accuracy comparable to the state-of-the-art.
\sysname, when used to augment vehicle perception, can improve safety.
When used in conjunction with offloaded planning, \sysname can increase traffic throughput at intersections.

\end{abstract}


\begin{IEEEkeywords}
Cooperative Perception, Infrastructure-assisted Perception, Autonomous Vehicle Systems
\end{IEEEkeywords}

\section{Introduction}
\label{sec:intro}

Machine perception extracts higher-level representations of a scene from low-level sensor signals in real-time.
Perception is essential for autonomy and is now a crucial component of every autonomous driving stack~\cite{BaiduSpecs,Autoware}.
The perception component of an autonomous driving stack extracts bounding boxes and tracks of dynamic objects in a scene such as vehicles, pedestrians and bicyclists.
It may also extract compact representations for static scene elements such as lane markers and drivable space.

\begin{figure}[t]
  \centering
  \includegraphics[width=0.65\columnwidth]
  {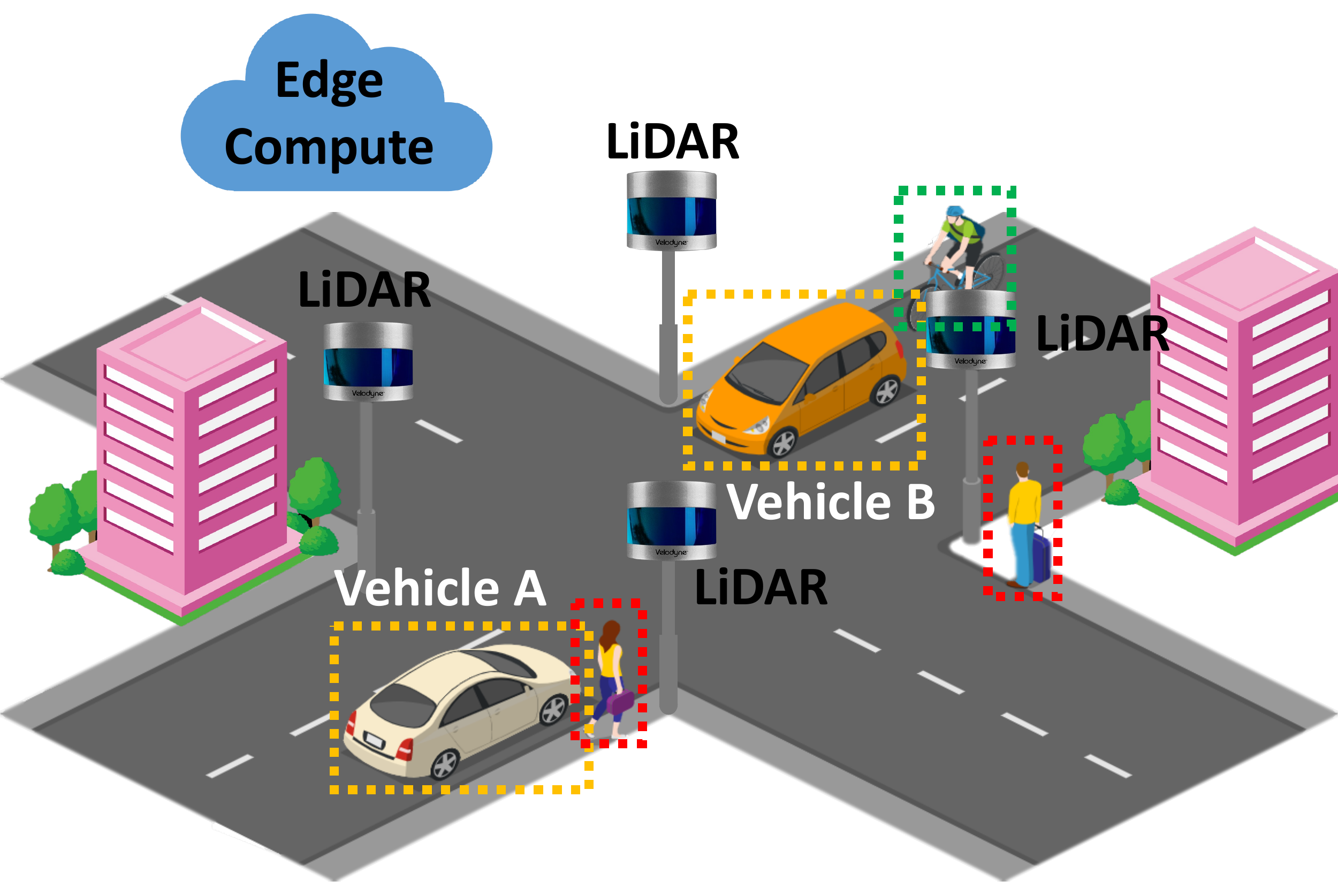}
  \caption{\sysname deployment at an intersection with multiple LiDARs and nearby edge compute.}
  \label{fig:deployment}
\end{figure}

Perception has long suffered from sensor range and line-of-sight limitations.
For example, a LiDAR on the vehicle A in \cref{fig:deployment} may not have enough range to see the bicyclist behind the vehicle B.
Even if it did, the LiDAR's view would be occluded by the vehicle B.
To address this, prior work has considered two qualitatively different approaches.
In \textit{cooperative perception}, vehicles share sensor or perception outputs between themselves~\cite{autocast} to effectively extend visual range and address line-of-sight limitations.
For example, the vehicle A (\cref{fig:deployment}) could ``see'' the bicyclist using vehicle B's perception outputs.
In \textit{infrastructure-assisted perception}~\cite{vips}, a vehicle augments its own perception using sensors in the infrastructure~\cite{ouster_chattanooga}.
In \cref{fig:deployment}, the vehicle A could ``see'' the occluded bicyclist using the LiDAR at the top of the figure.

In this work, we consider a complementary capability, \textit{cooperative infrastructure perception} (or \sysname).
This produces bounding boxes and tracks of dynamic objects in a scene, such as vehicles and pedestrians, by combining and processing outputs of multiple infrastructure sensors.
\cref{fig:deployment} shows an example in which \sysname can use four LiDARs placed at an intersection to cooperatively detect and track vehicles, pedestrians and bicyclists.
Compared to cooperative perception, \sysname's LiDARs, when mounted well above vehicle height, will likely have a better view of objects in the scene.
For example, the LiDAR on the left might be able to view part of the bicyclist behind the vehicle B, while the vehicle A's LiDAR may not.
Compared to infrastructure-assisted perception, \sysname's LiDARs can collectively obtain more complete views of an object and so can estimate better bounding boxes.
For example, each LiDAR adjacent to the vehicle B can only view half of the vehicle; together, they can view the entire car.

\sysname enables several novel capabilities:

\begin{description}[nosep,leftmargin=*]
\item[Perception Augmentation.] \sysname can deliver bounding boxes and tracks of all traffic participants to all vehicles.
An autonomous vehicle can augment its own perception with these and plan better paths, thereby increasing overall safety.

\item[Planning Offload.] \sysname's outputs can be used to plan trajectories for all vehicles on edge compute (\cref{fig:deployment}), and deliver each vehicle's trajectory wirelessly.
Offloading planning can be useful in restricted settings such as ports, parking lots, factories, and so on.
Indeed, industry is developing such a capability to autonomously guiding a car without on-board perception and planning into a parking spot~\cite{seoulrobotics}.

\item[Pedestrian Situational Awareness.] \sysname's outputs can be processed to deliver situational awareness to pedestrians such as an audio cue to a visually-impaired pedestrian or rendering by a real-time outdoor augmented reality system.
\end{description}

More generally, \sysname can be deployed not just at intersections, but in plazas, shopping malls, college and enterprise campuses and tourist attractions, and can be used to guide pedestrians, bicyclists and vehicles in various ways.

Three technology trends enable \sysname.
LiDARs are becoming cheaper, especially with the development of solid-state LiDARs~\cite{sslidars}.
Large cloud providers are rolling out edge compute deployments~\cite{GDCE}.
Finally, cellular carriers have made substantial investments in 5G deployments~\cite{Qualcomm5g}.

Given these trends, we expect that \sysname software can be architected as a software pipeline running on commodity edge compute (\cref{fig:deployment}).
\sysname will process and combine LiDAR outputs to produce objects and tracks, then deliver them wirelessly to traffic participants.
The use cases presented above motivate two challenging requirements that \sysname must meet.

\begin{itemize}[nosep,leftmargin=*]
\item Autonomous vehicles must perceive the world and make driving decisions at a frequency of 10~Hz with a tail latency less than 100~ms~\cite{lin18:_archit_implic_auton_drivin}.
To be applicable to autonomous driving, \sysname must also adhere to the same latency constraints.
LiDARs generate data at 10 frames per second.
At 30~MB per frame, with four LiDARs at an intersection, this translates to \sysname having to process data at a raw rate of nearly 10~Gbps.
With commodity compute, this is not straightforward.

\item \sysname generates a scene description that consists of dynamic objects along with their positions, bounding boxes, heading vectors and motion vectors.
The accuracy of these must match or exceed the state-of-the-art computer vision algorithms.
\end{itemize}

It is not immediately obvious that \sysname can meet these requirements; for example, the most accurate 2D and 3D object detectors on a popular autonomous driving benchmark~\cite{geiger2013vision} incur a processing latency of 60-300~ms.


To address these challenges, our work makes the following contributions:

\begin{itemize}[nosep,leftmargin=*]
\item To fuse 3D frames from multiple LiDARs, \sysname introduces a novel alignment algorithm whose accuracy is significantly higher than prior work (\cref{s:fusion}).

\item With an accurate 3D fused view, \sysname introduces fast and cheap implementations of algorithms for dynamic object extraction, tracking, and speed estimation.
These are centered around a single bounding box abstraction which \sysname computes early on in the pipeline (\cref{s:part-extr}).
This design choice is crucial for ensuring speed without sacrificing accuracy.

\item Cheap algorithms for heading estimation are inaccurate, so \sysname develops a more accurate GPU-offload heading estimator to meet the latency constraint (\cref{s:tracking}).

\item Its outputs can be used to augment vehicle perception, or enable offloaded planning, capabilities that can increase vehicular safety and throughput.
\end{itemize}

Using real-world datasets and simulations, we show that \sysname can generate perception outputs with a 99th percentile latency of less than 90~ms in scenes with 30-50 vehicles and pedestrians, just using a 16 core desktop with a single GPU.
Its object extraction and tracking accuracy compare well with the state-of-the-art.
When used to augment perception, it can ensure safety in 3$\times$-5$\times$ more scenarios than standard autonomous driving.
When used with offloaded planning, it can reduce traffic wait time by up to 5$\times$.

\section{\sysname Design}
\label{sec:design}

\begin{figure*}[t]
  \centering
  \begin{minipage}{0.19\linewidth}
    \centering
    \includegraphics[width=\columnwidth,frame]
    {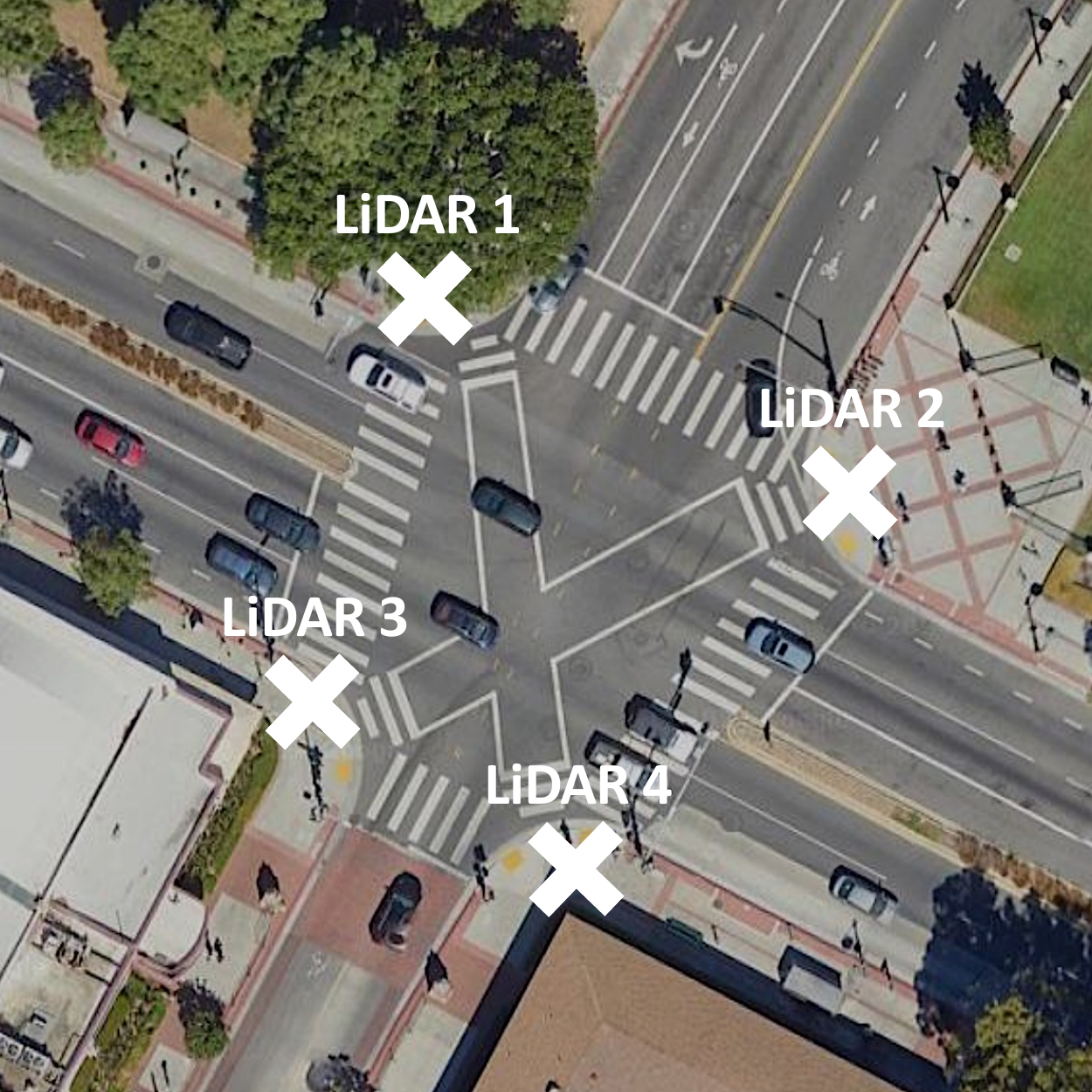}
    {\small (a)}
  \end{minipage}
  \begin{minipage}{0.19\linewidth}
    \centering
    \includegraphics[width=\columnwidth,frame]{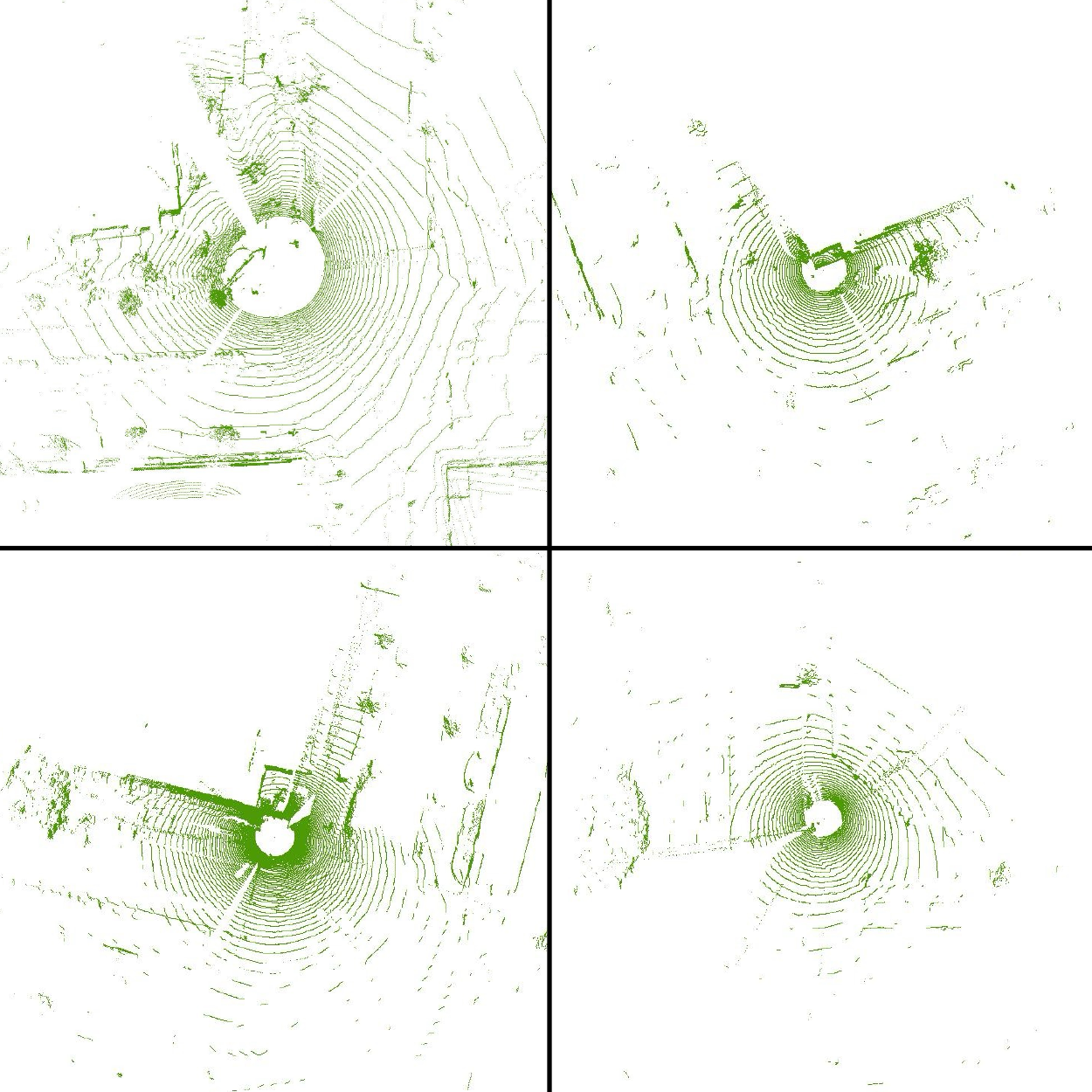}
    {\small (b)}
  \end{minipage}
  \begin{minipage}{0.19\linewidth}
    \centering
    \includegraphics[width=\columnwidth,frame]{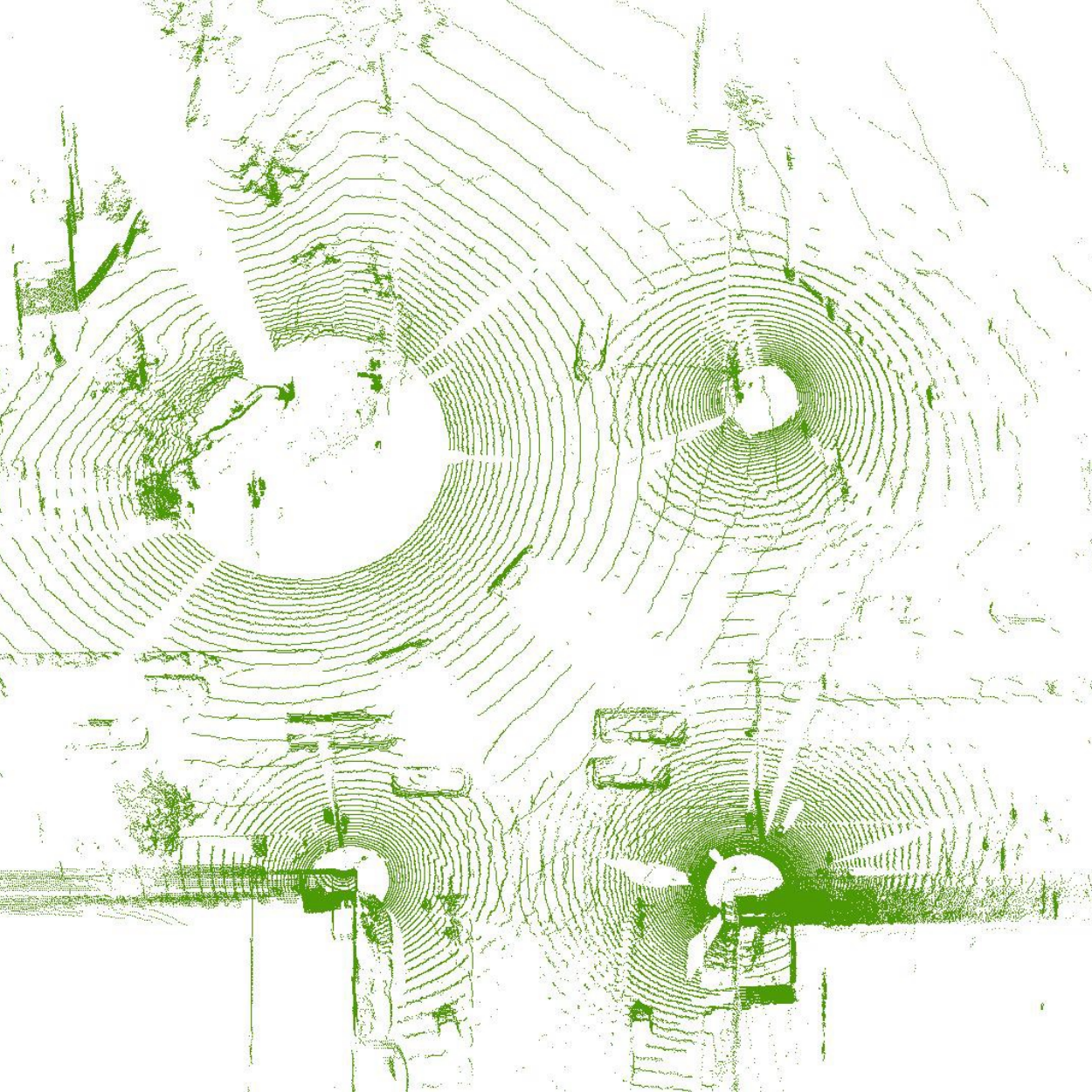}
    {\small (c)}
  \end{minipage}
  \begin{minipage}{0.19\linewidth}
    \centering
    \includegraphics[width=\columnwidth,frame]{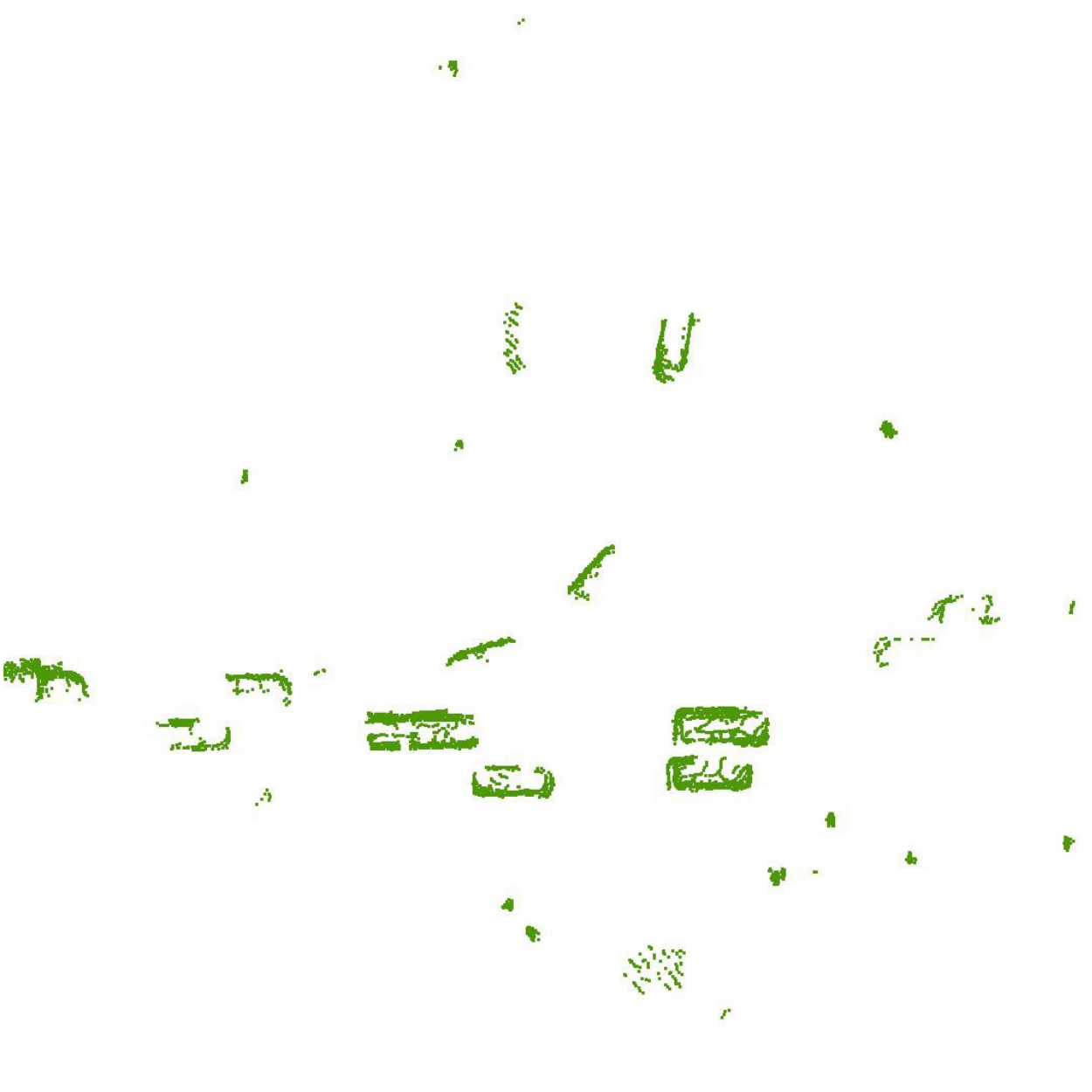}
    {\small (d)}
  \end{minipage}
  \begin{minipage}{0.19\linewidth}
    \centering
    \includegraphics[width=\columnwidth,frame]{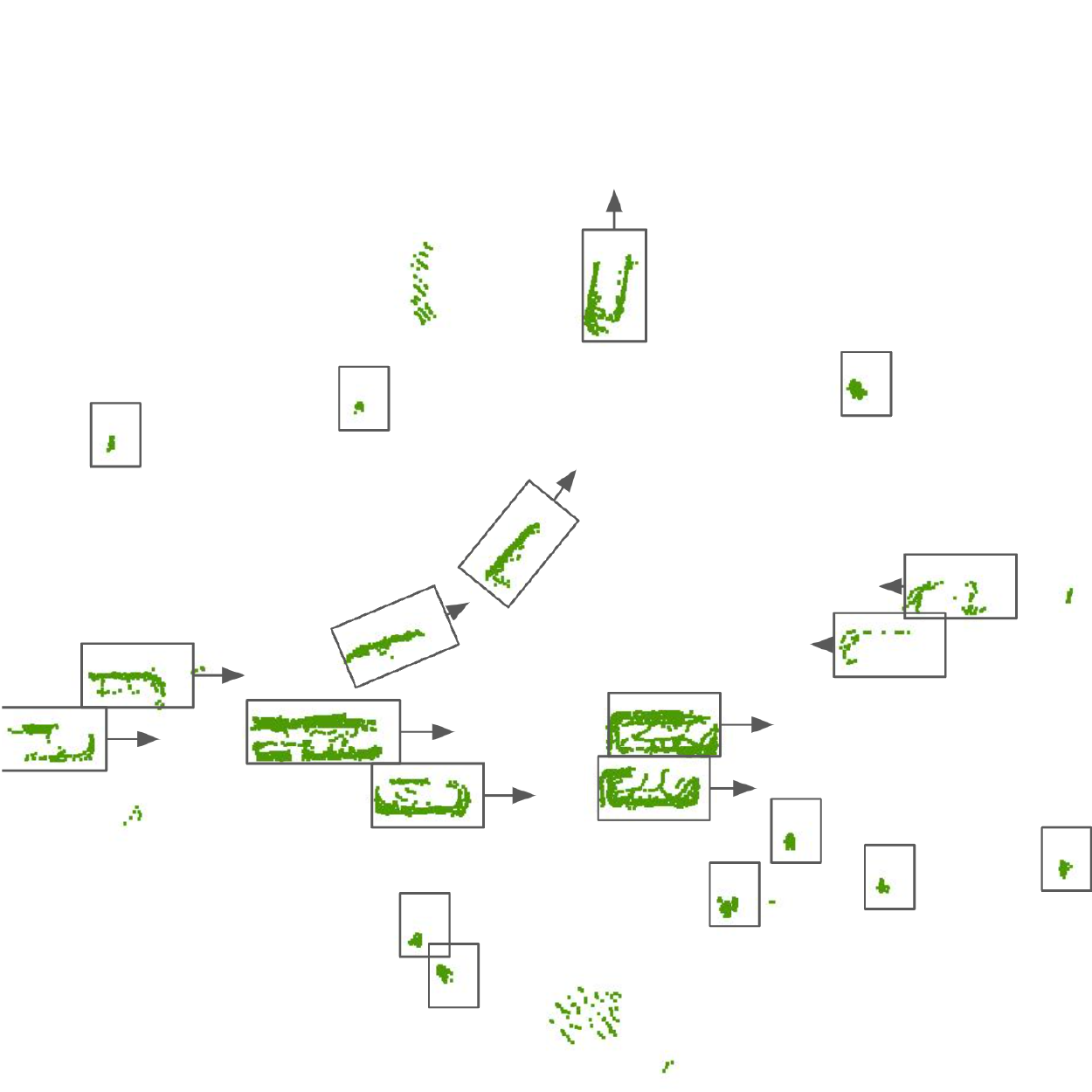}
    {\small (e)}
  \end{minipage}
  \caption{\sysname deployment at a busy intersection with heavy vehicular and pedestrian traffic in a large metropolitan city. (a) A top down view of the intersection (taken from Google Maps~\cite{google_maps}). We mounted four LiDARs near each of traffic light poles situated at the four corners of the intersection. (b) An individual frame from each one of the four LiDARs. (c) A fused frame. (d) Point clouds of traffic participants (dynamic objects) at the intersection. (e) Bounding boxes and motion vectors for traffic participants, calculated over successive frames.}
  \label{fig:intersection_example}
\end{figure*}


\begin{figure}[t]
  \centering
  \includegraphics[width=0.8\columnwidth]{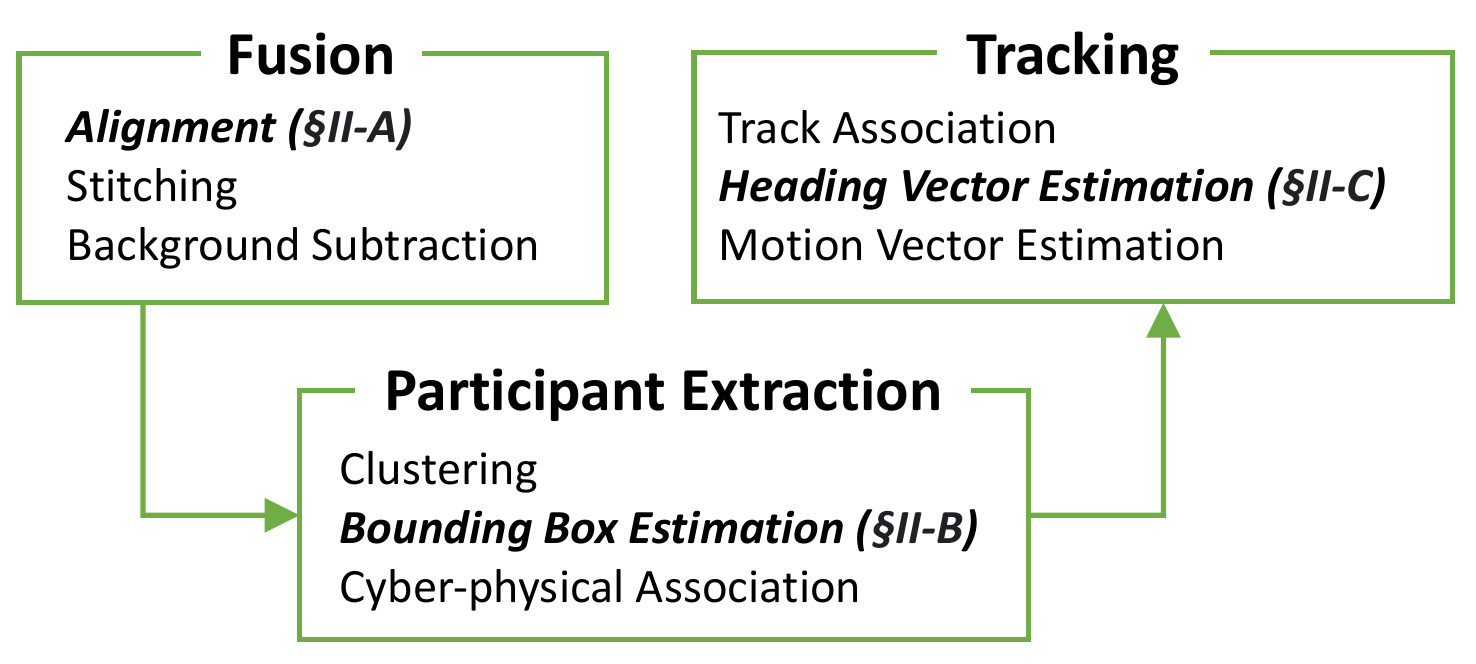}
  \caption{Perception stages. Bold sub-stages described in detail.} 
  \label{fig:perception_stages}
\end{figure} 

In this section, we describe \sysname's design, beginning with an overview of its approach.
%
%
We do this using data from our deployment of four LiDARs at the four corners of a busy intersection of a major metropolitan area as shown in \cref{fig:intersection_example}(a).\footnote{Practical deployments of roadside LiDARs need to consider coverage redundancy and other placement geometry issues, which are beyond the scope of this paper.}

\parab{Inputs and Outputs.}
The input to \sysname is a continuous sequence of LiDAR frames from each LiDAR in a set of \textit{overlapping} LiDARs deployed roadside.
The output of perception is a compact \textit{abstract scene description}: a list of bounding boxes of moving objects together with their motion vectors (\cref{fig:intersection_example}(e)).

\begin{figure*}
  \centering
  \begin{minipage}{0.20\linewidth}
    \centering
    \includegraphics[width=\columnwidth,frame]{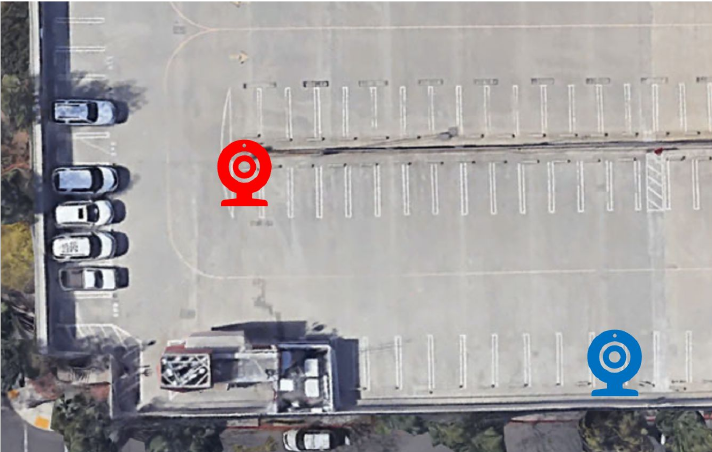}
    {\small (a)}
  \end{minipage}
  \begin{minipage}{0.20\linewidth}
    \centering
    \includegraphics[width=\columnwidth,frame]{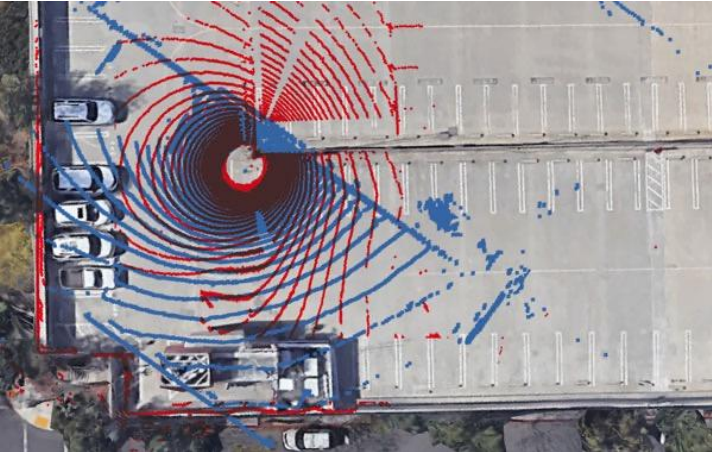}
    {\small (b)}
  \end{minipage}
  \begin{minipage}{0.15\linewidth}
    \centering
    \includegraphics[width=0.9\columnwidth,frame]{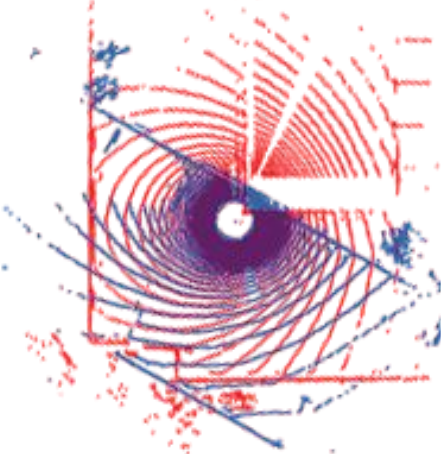}
    {\small (c)}
  \end{minipage}
  \begin{minipage}{0.1725\linewidth}
    \centering
    \includegraphics[width=0.9\columnwidth,frame]{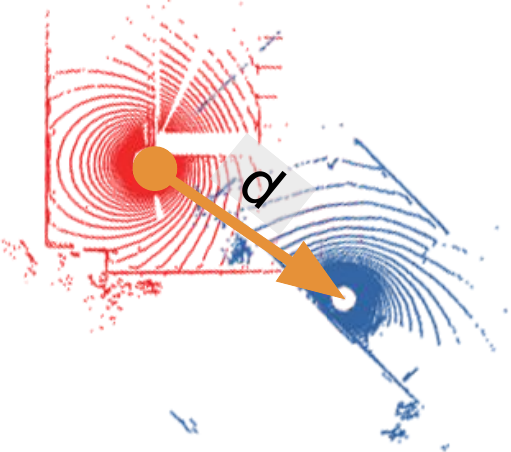}
    {\small (d)}
  \end{minipage}
  \begin{minipage}{0.24\linewidth}
    \centering
    \includegraphics[width=0.9\columnwidth,frame]{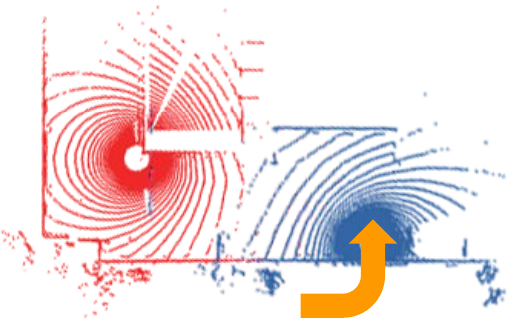}
    {\small (e)}
  \end{minipage}
  \caption{An illustration of \sysname's point cloud alignment algorithm. (a) A top down view of a parking lot with two LiDARs shown by red ($L_{1}$) and blue ($L_{2}$) icons. (b) The inputs to initial guess algorithm are point clouds ($C_{1}$ and $C_{2}$) in the respective LiDAR's coordinate system along with the ground distance $d$ between them. (c) Figure (b) with background removed. (d) To fix the base coordinates, \sysname displaces $C_{2}$ by the ground distance $d$. (e) \sysname rotates both $C_{1}$ and $C_{2}$ by small yaw increments to find the combination with the least distance between the point clouds.}
  \label{fig:alignment}
\end{figure*}


\parab{Approach.}
To address the challenges described in \cref{sec:intro}, \sysname uses a three stage pipeline, each with three sub-stages (\cref{fig:perception_stages}).
\textit{Fusion} combines multiple LiDAR views into fused frames (\cref{fig:intersection_example}(c)), and then subtracts the static background to reduce data for subsequent processing (\cref{fig:intersection_example}(d)).
\textit{Participant extraction} identifies traffic participants by clustering and, estimates a tight 3D bounding box around each object.
\textit{Tracking} associates objects across frames, and estimates their heading and motion vectors (\cref{fig:intersection_example}(e)).

This design achieves \sysname's goals using three ideas:

\begin{itemize}[nosep,wide]
\item \sysname exploits the fact that LiDARs are static to cheaply \textit{fuse} point clouds from multiple overlapping LiDARs into a single \textit{fused frame}.
Such a frame may have more complete representations of objects than individual LiDAR frames.
\cref{fig:intersection_example}(b) shows the view from each of the four LiDARs; in many of them, only parts of a vehicle are visible.
\cref{fig:intersection_example}(c) shows a \textit{fused frame} which combines the four LiDAR frames into one; in this, all vehicles are completely visible.\footnote{Not immediately obvious from the figure, but a LiDAR frame, or a fused frame is a 3D object, of which \cref{fig:intersection_example}(c) is a 2D projection.}

\item \sysname builds most algorithms around a single abstraction, the 3D bounding box of an object (\cref{fig:intersection_example}(e)).
Its tracking, speed estimation, and motion vector estimation rely on the observation that the centroid of the bounding box is a convenient consistent point 
for estimating these quantities, especially when fused LiDAR frames provide comprehensive views of an object.

\item \sysname uses, when possible, cheap algorithms rather than expensive deep neural networks (DNNs).
Only when higher accuracy is required does \sysname resort to expensive algorithms, but employs hardware acceleration to meet the latency budget; its use of a specialized heading vector estimation 
is an example.
\end{itemize}

To appreciate the novelty of this approach, consider \cref{tab:stack_comparison} which compares \sysname's design to that of open-source autonomous driving perception designs~\cite{BaiduSpecs,Autoware}:

\begin{itemize}[nosep,wide]
\item These approaches 
rely on a pre-built 3D map (called HD map), and localize the ego-vehicle (the one on which the autonomous driving stack runs) by matching its LiDAR scans against the HD map.
In contrast, \sysname does not require a map for positioning: all vehicle positions can be directly estimated from the fused frame.

\item Autonomous driving stacks use different ways to identify traffic participants.
They extract obstacles from a 2D birds-eye-view or use a 2D object detector on images, then back-project these to the 3D LiDAR view.
\sysname, on the other hand, directly extracts the 3D point cloud associated with each participant.
It does this cheaply using background subtraction because its LiDARs are relatively static.

\item Autonomous driving stacks use Kalman filters to estimate motion properties\footnote{Vehicles use SLAM to estimate their own motion and heading.}
(\eg speed and heading) of other vehicles.
For heading, \sysname uses a more sophisticated algorithm to ensure higher tail accuracy.
\end{itemize}

Below, we describe parts of \sysname's novel perception relative to autonomous driving stacks (bold text in \cref{tab:stack_comparison}). 

\begin{table}[t]
\scriptsize
\centering
\begin{tabular}{p{0.65in}|p{1in}|p{1in}} \hline
                               & \textbf{\sysname} & \textbf{AV Perception}~\cite{BaiduSpecs,Autoware}  \\ \hline\hline
Mapping                        & \textit{\textbf{None}} & Pre-built HD map      \\ \hline
Localization                   & \textit{\textbf{Live Fused Point Cloud}} (\cref{s:fusion}) & LiDAR scan matching, Fusion with GPS-RTK or IMU   \\ \hline
Object Detection               & \textit{\textbf{3D Object Detection}} (\cref{s:part-extr}) & Obstacle detection in 2D BEV projection, 2D object detection in camera, projected to 3D \\ \hline
Tracking/Motion Estimation & Kalman filter with matching, \textit{\textbf{Heading Estimation}} (\cref{s:tracking}) & Kalman filter with matching      \\ \hline
\end{tabular}
\caption{\sysname and autonomous vehicle perception.}
\label{tab:stack_comparison}
\end{table}

\subsection{Accurate Alignment for Fast Fusion}
\label{s:fusion}

\parab{Point Cloud Alignment.}
Each frame of a LiDAR contains a \textit{point cloud}, a collection of points with 3D coordinates.
These coordinates are in the LiDAR's own frame of reference.
Fusing frames from two different LiDARs is the process of converting all 3D coordinates of both point clouds into a common frame of reference.
\textit{Alignment} computes the transformation matrix for this conversion.

Prior work has developed Iterative Closest Point (ICP)
~\cite{chen91:_objec} techniques that \textit{search} for the lowest error alignment.
The effectiveness of these approaches depends upon the initial guess for LiDARs' poses.
Poor initial guesses can result in local minima.
SAC-IA~\cite{rusu09:_fast_point_featur_histog_fpfh} is a well-known algorithm to quickly obtain an initial guess for ICP.
As we demonstrate in \cref{sec:evaluation}, with SAC-IA's initial guesses, ICP generates poor alignments on full LiDAR frames.




\parab{Initial Guess using Minimal Information.}
Besides the point clouds, SAC-IA requires no additional input.
\sysname uses an algorithm to obtain good initial guesses using minimal additional input.
%
%
Specifically, \sysname only needs the distances on the ground between a reference LiDAR and all others to get good initial guesses.\footnote{LiDAR GPS locations as input result in poor alignment (\cref{sec:evaluation}).}
We can measure these distances using, for example, an off-the-shelf laser rangefinder~\cite{duchon2012some}.

We now describe the algorithm for two LiDARs $L_1$ and $L_2$ (as shown in \cref{fig:alignment}(a)); the technique generalizes to multiple LiDARs as described in \cref{algo:icp}.
The inputs are two point clouds $C_1$ and $C_2$ (more generally, $N$ point clouds captured from the corresponding LiDARs at the same instant (\cref{fig:alignment}(b) and \cref{fig:alignment}(c)), and the distance $d$ on the ground between the LiDARs.
The output is an initial guess for the pose of each LiDAR.
We feed these guesses into ICP to obtain good alignments.
The algorithm conceptually consists of three steps:

\parae{Fix base coordinates.}
Set $L_1$'s $x$ and $y$ coordinates to be $(0,0)$ \ie base at origin (\cref{algo:icp} line 3).
Then, assume that $L_2$'s base is at $(d,0)$ (\cref{fig:alignment}(d)).

\parae{Estimate height, roll and pitch.}
In this step, we determine: the height of each LiDAR $z_i$, the roll (angle around the $x$ axis), and pitch (angle around the $y$ axis).
For these, \sysname relies on fast \textit{plane-finding} algorithms~\cite{derpanis2010overview} that extract planes (\cref{algo:icp} lines 1 and 4) from a collection of points.
These techniques output the equations of the planes.
Assuming that the largest plane is the \textit{ground-plane} (a reasonable assumption for roadside LiDARs), \sysname aligns the $z$ axis of two LiDARs with the normal to the ground plane (\cref{algo:icp} line 5).
In this way, it implicitly fixes the roll and pitch of the LiDAR.
Moreover, after the alignment, the height of the LiDARs $z_i$ is also known (because the $z$ axis is perpendicular to the ground plane).

\parae{Estimate yaw.} 
Finally, to determine yaw (angle around the $z$ axis), we use the technique illustrated in \cref{fig:alignment}(e) (\cref{algo:icp} line 6).
In this technique, \sysname rotates both point clouds $C_{1}$ and $C_{2}$ with different yaw settings until it finds a combination that results in the smallest 3D distance\footnote{The 3D distance between two point clouds is the average distance between every point in the first point cloud to its nearest neighboring point in the second point cloud.} between the two point clouds.
We have found that ICP is robust to initial guesses for yaw that are within about 15-20$^\circ$ of the actual yaw, so \sysname discretizes the search space by this amount.
In case of \cref{fig:alignment}(e), \sysname calculates the initial guess for the yaw by rotating the blue point cloud ($C_{2}$).

\parab{Obtaining Alignment.}
\sysname repeats this procedure for every other LiDAR $L_i$ with respect to $L_1$, to obtain initial guesses for the poses of every LiDAR (\cref{algo:icp} line 2).
It feeds these into ICP to obtain an alignment.

Alignment is run only once when installing the LiDARs.
Re-alignment may be necessary if a LiDAR is replaced or re-positioned.
Alignment is performed infrequently but is crucial to \sysname's accuracy (\cref{sec:eval-microbenchmark}).

\parab{Per-frame Stitching.}
LiDARs generate frames at 10~fps (or more).
In \cref{fig:intersection_example}, when each LiDAR generates a frame, the fusion stage performs \textit{stitching}.
Stitching applies the coordinate transformation for each LiDAR generated by the alignment resulting in a fused frame (\cref{fig:intersection_example}(c)).

\begin{algorithm}[t]
\small
\SetAlgoLined
\SetKwInOut{Input}{Input}
\SetKwInOut{Output}{Output}
\SetKw{KwBy}{by}

\Input{1. $\mathbf{C} = \{C_1,  \cdots, C_N\}$; $C_i$ is LiDAR $L_i$'s point cloud. 
\newline
 2. $D = \{d_2, \cdots, d_N\}$; $d_i$ is the distance from $L_i$ to the reference LiDAR $L_1$.

}
\Output{$T = \{T_2, T_3, \cdots, T_N\}$; $T_i$ is the transformation matrix from $C_i$ to $L_1$'s coordinate system.}

SegmentGroundPlane ( $C_1$ )\label{icp:ln1}\;

\For{$i\gets2$ \KwTo $N$ \KwBy $1$}{

  AlignPosition ( $C_i$, $C_1$, $d_i$ )\label{icp:ln5}\;

  SegmentGroundPlane ( $C_i$ )\label{icp:ln3}\;

  AlignGroundPlane ( $C_i$, $C_1$ )\label{icp:ln4}\;

  EstimateYaw ( $C_i$ )\label{icp:ln6}\;

}
\caption{Estimating an initial guess for alignment.}
\label{algo:icp}
\end{algorithm}

\subsection{Reusing 3D Bounding Boxes}
\label{s:part-extr}

\sysname's efficiency results from reusing the \textit{3D bounding box} of a participant (\cref{fig:intersection_example}(e)) in processing steps.
After stitching, \sysname extracts the bounding box by performing \textit{background subtraction} on the fused point cloud to extract points belonging to dynamic objects.
On these points, it applies \textit{clustering} to determine points belonging to individual objects.
Finally, it runs a \textit{bounding box estimation} algorithm on these points.
These use well-known algorithms, albeit with some optimizations; we describe these in \cref{s:optimizations}.
\sysname uses the bounding box for many of its algorithms (\cref{tab:bounding_boxes}); we describe these below.


%

\begin{table}[t]
\small
\begin{tabular}{ccc|c|c}
\hline
  \multicolumn{3}{c|}{\textbf{Centroid}} &
  \textbf{Axes} &
  \textbf{Dimension} \\ \hline \hline
  \multicolumn{1}{c|}{Tracking} &
  \multicolumn{1}{c|}{\begin{tabular}[c]{@{}c@{}}Speed \\ Estimation\end{tabular}} &
  \begin{tabular}[c]{@{}c@{}}Cyber-phy\\ Association\end{tabular} &
  \begin{tabular}[c]{@{}c@{}}Heading \\ Estimation\end{tabular} &
  Planning \\ \hline
\end{tabular}
\caption{\sysname reuses properties of the bounding boxes in multiple modules to ensure low latency.}
\label{tab:bounding_boxes}
\end{table}


\parab{Tracking.}
To associate objects across frames (\textit{tracking}), \sysname uses a Kalman filter to predict the position of the \textit{centroid} of the 3D bounding box.
Then, it finds the best match (in a least-squares sense) between predicted positions and the actual positions of the centroids in the frame.
Although tracking in point clouds is a challenging problem~\cite{vaquero2017deconvolutional} for which research is exploring deep learning, a Kalman filter works \textit{exceedingly well} in our setting.
The biggest challenge in tracking is occlusions: when one object occludes another in a frame, it may be mistaken for the other in subsequent frames (\textit{ID-switch}).
Because our fused frame includes perspectives from multiple LiDARs, ID-switches occur rarely (\cref{sec:evaluation}).

\parab{Speed Estimation.}
To estimate \textit{speed} of a dynamic object, \sysname measures the distance between the centroid of the bounding box in one frame and the centroid $w$ frames in the past ($w$ is a configurable window size).
It then estimates speed by dividing the distance by the time to generate $w$ frames.

\parab{Cyber-physical Association.} 
\sysname needs to associate an object seen in the LiDAR with a cyber endpoint (\eg an IP address).
This is important so that \sysname can send that object customized results \ie perception results relevant to that object or a customized trajectory planned for that vehicle.
For this step, 
\sysname uses a calibration step performed once.
%
Given a vehicle for which we know the cyber-physical association (\eg LiDAR installer's vehicle), we estimate the transformation between the trajectory of the vehicle seen in the LiDAR view with the GPS trajectory (details omitted for brevity).
\sysname uses this to transform a vehicle's GPS trajectory to its expected trajectory in the scene, then matches actual scene trajectory to expected trajectory in a least-squares sense.
To define the scene trajectory, we use the centroid of the vehicle's bounding box.



The centroid of the bounding box is an easily computed and consistent point within the vehicle that simplifies these tasks.
Because we have multiple LiDARs that capture a vehicle from multiple directions, the centroid of the bounding box is generally a good estimate of the actual centroid of the vehicle.

Besides these, \sysname (a) estimates heading direction from the axes of the bounding box (discussed in \cref{s:tracking}) and (b) uses the dimensions of the box to represent spatial constraints for planning (discussed in \cref{s:use_cases}). 

\subsection{Fast, Accurate Heading Vectors}
\label{s:tracking}

To compute the \textit{motion vector} of a vehicle, \sysname first determines, for each object, its instantaneous \textit{heading} (direction of motion), which is one of the three surface normals of the bounding box of the vehicle.
It estimates the motion vector as the average of the heading vectors in a sliding window of $w$ frames.
Most autonomous driving stacks can obtain heading from SLAM or visual odometry (\cref{sec:related_work}), so little prior work has explored extracting heading from infrastructure LiDAR frames.


\begin{figure}[t]
  \centering
  \includegraphics[width=0.7\columnwidth,frame]{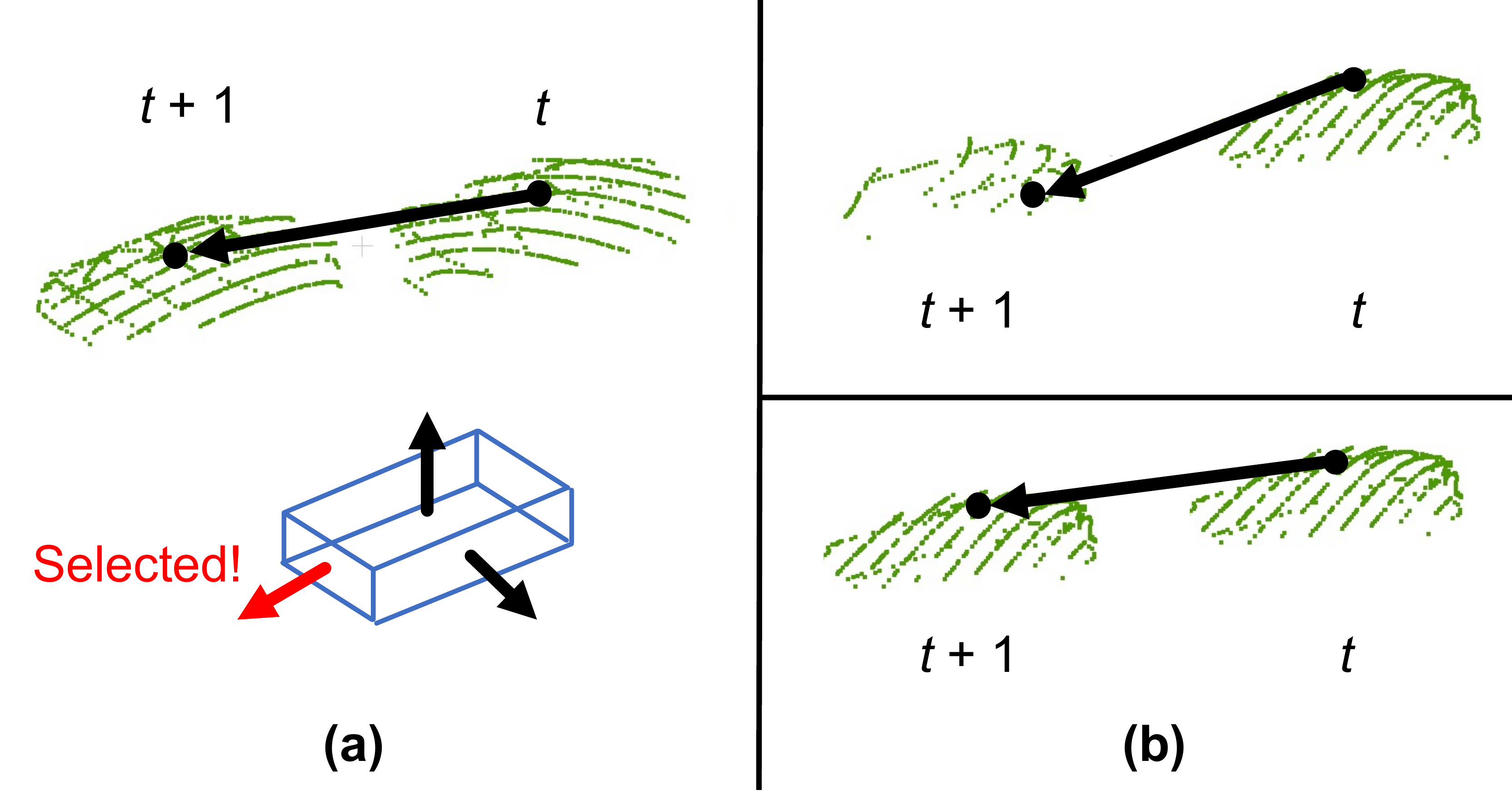}
  \caption{The figure shows the points belonging to a vehicle in two successive frames $t$ and $t+1$. (a) Strawman approach for heading determination. (b) \sysname's approach.}
  \label{fig:heading}
\end{figure}

\parab{A Strawman Approach.} 
Consider an object $A$ at time $t$ and time $t+1$.
\cref{fig:heading}(a) shows the points belonging to that object.
Since those points are already in the same frame of reference, a strawman algorithm finds the vector between the centroid of $A$ at time $t$ and centroid of $A$ at time $t+1$.
Then, the heading direction is the surface normal from the bounding box that is most closely aligned with this vector (\cref{fig:heading}(a)).
We have found that the error distribution of this approach can have a long tail (although average error is reasonable).
If $A$ has fewer points in $t+1$ than in $t$ (\cref{fig:heading}(b), upper), the computed centroid will be different from the true centroid, which can induce significant error.

\parab{\sysname's Approach.}
To overcome this, \sysname uses ICP to find the transformation matrix between $A$'s point cloud in $t$ and in $t+1$.
Then, it places $A$'s point cloud from $t$ in frame $t+1$ (\cref{fig:heading}(b), lower).
Finally, it computes the vector between the centroids of these two (so that the centroid calculations are based on the same set of points).
As before, the heading direction is the surface normal 
that is most closely aligned with this vector.

\parab{GPU Acceleration.} 
ICP is compute-intensive even for small object point clouds.
If there are multiple objects in the frame, \sysname must run ICP for each of them.
We have experimentally found this step to be the bottleneck.
Thus, we developed a fast GPU-based implementation of heading vector estimation, which reduces the overhead of this stage (\cref{s:simul-results:-laten}).

A typical ICP implementation has four steps: (1) estimating correspondence between the two input point clouds, (2) estimating transformation between the two point clouds, (3) applying the transformation to the source point cloud and (4) checking for ICP convergence.
The first step requires a nearest neighbor search; instead of using octrees, we adapt a parallelizable version described in~\cite{garcia2008fast} but use CUDA's parallel scanning to find the nearest neighbor.
The second step shuffles points in the source point cloud then applies the Umeyama algorithm~\cite{umeyama} for the transformation matrix.
We re-implemented this algorithm using CUDA's demean kernel and a fast SVD implementation~\cite{gao2018gpu}.
For the third and fourth steps, we developed custom CUDA kernels.
This step scales linearly with the number of vehicles but parallelizes easily to multiple GPUs; at intersections with many vehicles, \sysname can use edge computing resources with multiple GPUs.

\subsection{Optimizations}
\label{s:optimizations}

\parab{Background Subtraction.} 
\sysname removes points belonging to static parts of the scene\footnote{Static parts of the scene (e.g., an object on the drivable surface) might be important for path planning.
\sysname uses the static background point cloud to determine the drivable surface; this is an input to the planner (\secref{s:centralized_planning}).
We omit this for brevity.}.
This is: (a) especially crucial for voluminous LiDAR data, and (b) feasible in our setting because LiDARs are static.
It requires a calibration step to extract a \textit{background point cloud} from each LiDAR~\cite{kashani2015review}, then creates a \textit{background fused frame} using the results from alignment.
To extract the background point cloud, \sysname takes the intersection of a few successive point clouds and the aggregating intersections taken at a few different time intervals.


\parab{Subtraction before Stitching.}
\sysname can subtract the background fused frame from each fused frame generated by stitching. 
We have found that removing the background from each LiDAR frame (using its background point cloud), and then stitching points in the residual point clouds can significantly reduce latency.
Stitching scales with the number of points, which this optimization reduces significantly.

\parab{Leveraging LiDAR Characteristics.}
Many LiDAR devices only output \textit{returns} from reflected laser beams.
Generic background subtraction algorithm requires a nearest-neighbor search to match a return with the corresponding return on the background point cloud.
Some LiDARs (like Ouster~\cite{Ouster}), however, indicate \textit{non-returns} as well, so that the point cloud contains the output of every beam of the LiDAR.
For these, it is possible to achieve fast background subtraction by comparing corresponding beam outputs in a point cloud and the background point cloud.



\parab{Computing 3D Bounding Boxes.}
On the points in the fused frame remaining after background subtraction, \sysname uses a standard clustering algorithm (DBSCAN)~\cite{ester1996density} to extract multiple clusters of points where each cluster represents one traffic participant.
Then, it uses an off-the-shelf algorithm~\cite{pclbbox}, which determines a minimum oriented bounding box of a cluster using principal component analysis (PCA).
From these, it extracts the three surface normals of the object: the vertical axis, the axis in the direction of motion, and the lateral axis.

\section{Use Cases}
\label{s:use_cases}

Beyond describing \sysname, it is important to demonstrate its utility.
To this end, we describe its use in (a) augmenting vehicle perception and (b) offloading planning to the edge.
These improve traffic safety and throughput respectively (\cref{sec:evaluation}).


\subsection{Augmenting a Vehicle's Perception}
\label{s:augment_vehicle_perception}

\begin{figure}[t]
\begin{minipage}{0.49\linewidth}
    \centering
 \includegraphics[width=0.85\columnwidth,frame]{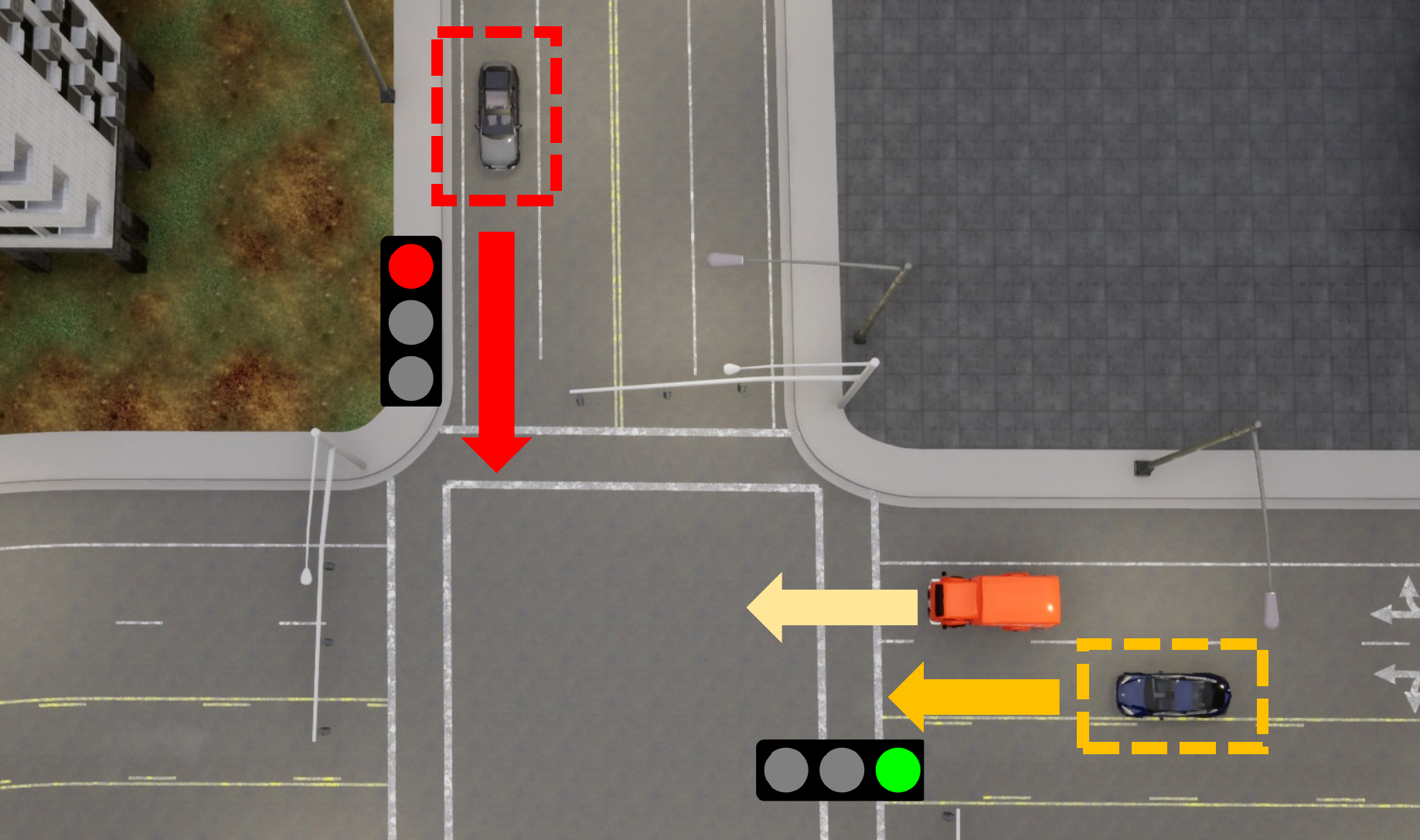}
    \caption{The orange truck obstructs ego-vehicle's (yellow box) view of the red-light violator.}
    \label{fig:red_light_violation}
\end{minipage}
\begin{minipage}{0.49\linewidth}
\centering
 \includegraphics[width=0.85\columnwidth,frame]{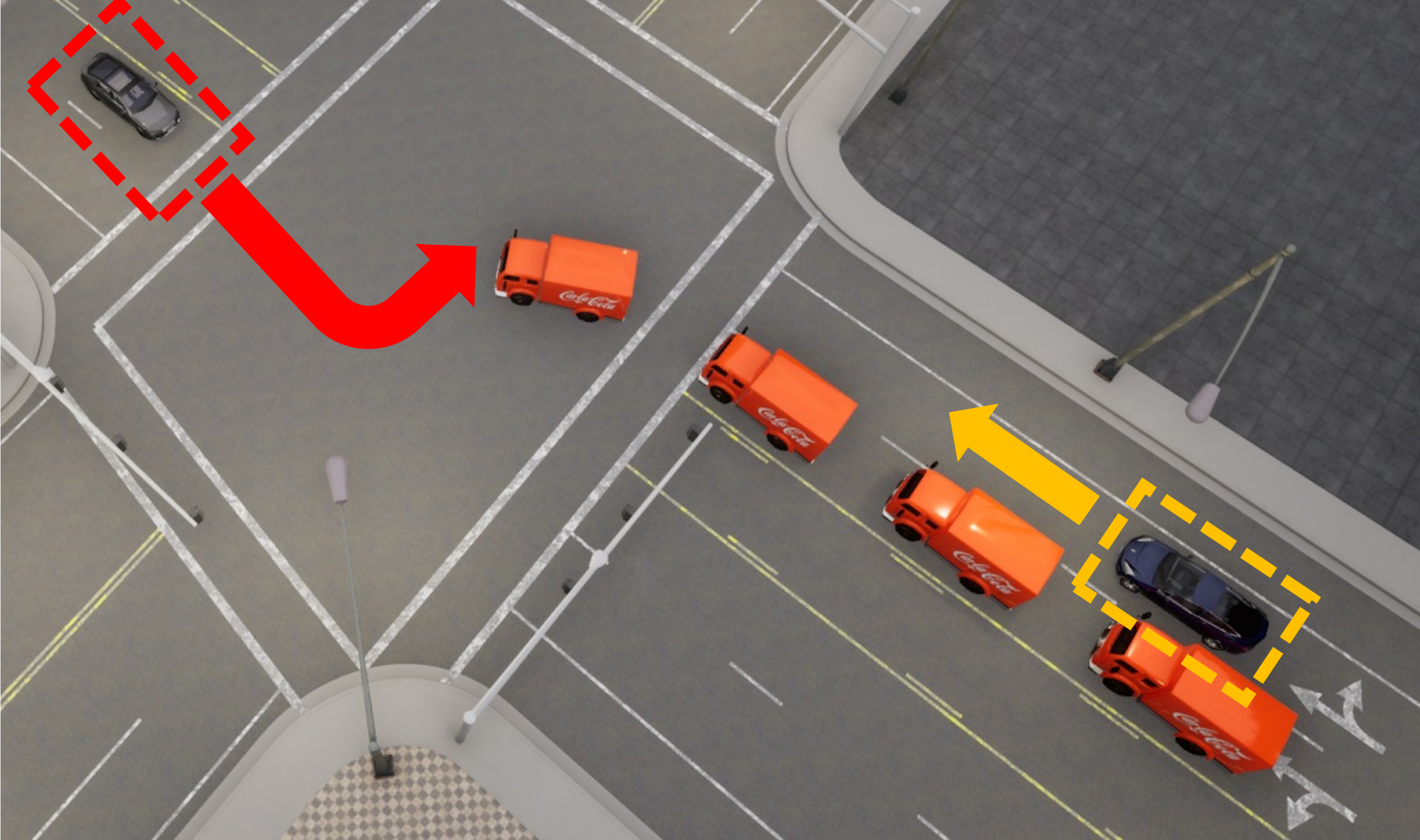}
    \caption{The orange trucks obstruct the ego-vehicle's (yellow box) view of the left-turning car (red box).}
    \label{fig:unprotected_left_turn}
\end{minipage}
\end{figure}

3D sensors mounted on autonomous vehicles are prone to line-of-sight-limitations and occlusions.
NHTSA~\cite{NHTSA-precrash} has highlighted a number of scenarios where line-of-sight limitations can cause potential traffic accidents~\cite{carlachallenge}.
For instance, in \cref{fig:red_light_violation} the ego-vehicle (bounded in yellow) has the right-of-way and attempts to cross the intersection.
However, it is unaware of an oncoming red-light violating vehicle (bounded in red) which is occluded by the orange truck.
As a result, the ego-vehicle will crash into the red-light violating vehicle.
The same is true for \cref{fig:unprotected_left_turn} where the ego-vehicle (bounded in yellow) cannot see the vehicle taking the unprotected left turn (bounded in red).



In such traffic scenarios, if the vehicle had access to other 3D views that could sense the oncoming vehicle, it could avoid the traffic accident.
With LiDARs mounted at the intersection, \sysname can wirelessly transmit
its perception outputs to all vehicles.
Vehicles, after positioning these bounding boxes and motion vectors in their own coordinate system~\cite{vrf}, can fuse them with results from their on-board perception stack.

Finally, they feed the fused results to their on-board planning module.
In an autonomous vehicle, the planner~\cite{paden2016survey} generates, every LiDAR frame, short-term \textit{trajectories} (at the timescale of 100s of milliseconds) that the vehicle must follow.
A trajectory is a sequence of way-points, together with the precise times at which the vehicle must arrive at those way-points.
Because the fused perception results contain the obstructed vehicles, the  planner is aware of their presence and their motion.
As a result, it can devise collision-free trajectories for vehicles to enable safer driving.

\subsection{Offloading Planning to the Edge}
\label{s:centralized_planning}

In \cref{s:augment_vehicle_perception}, each vehicle plans its trajectory independently.
However, because \sysname has a comprehensive view of the intersection, it can actually plan trajectories for \textit{all} vehicles on edge compute.
While this may seem far-fetched, there already exists at least one company~\cite{seoulrobotics} exploring this capability in limited settings.
In malls and airports, this capability uses infrastructure sensors and edge-based planning to guide vehicles to and out of their parking spaces.

Offloading planning to the edge can improve traffic throughput at intersections.
Instead of traffic lights, the planner can regulate the speed of each car so that all cars can traverse the intersection safely, possibly without stopping.
Traffic-light free intersections~\cite{traffic-light-free} are a long sought after goal in the transportation literature.


\begin{figure}[t]
  \centering
  \begin{minipage}{0.325\linewidth}
    \centering
    \includegraphics[width=\columnwidth,frame]{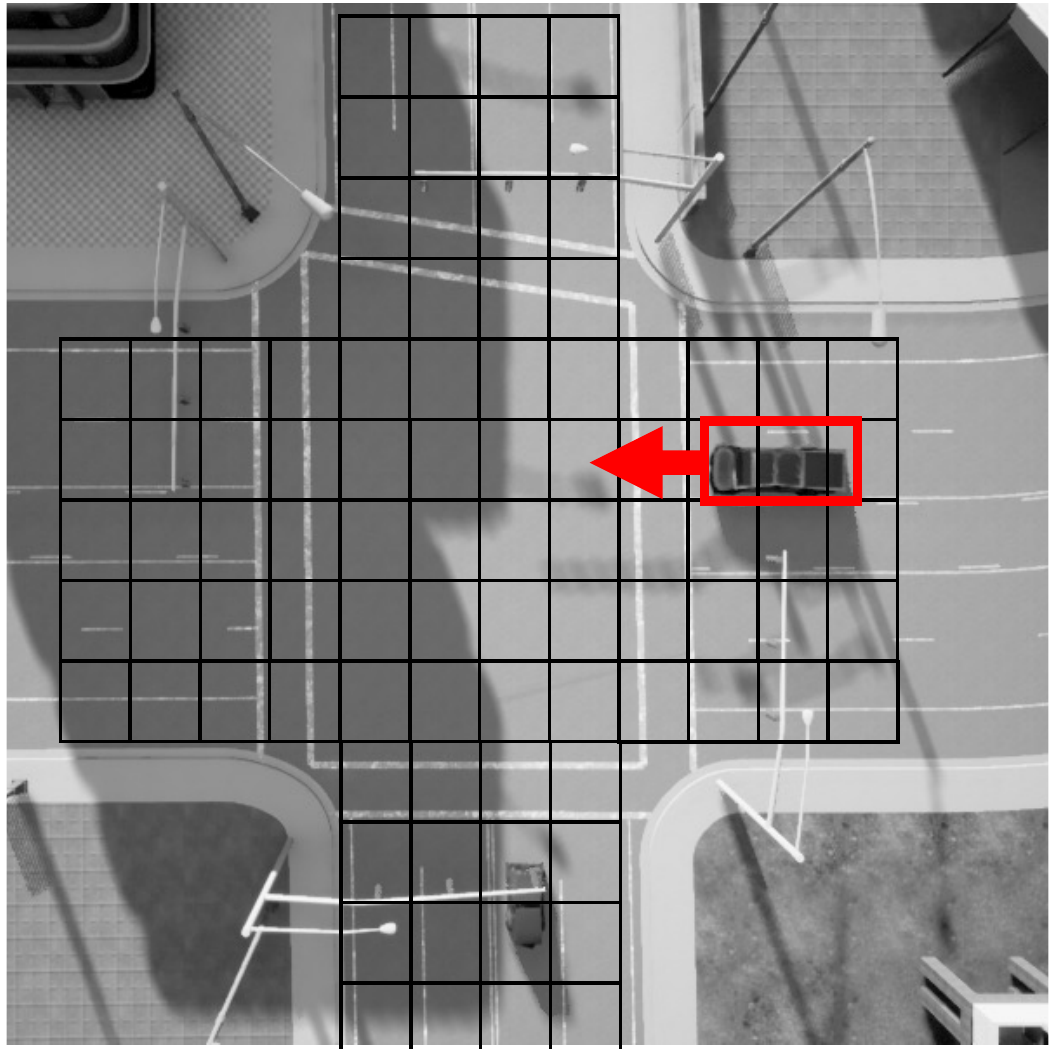}
    \\ {\small (a)}
  \end{minipage}
  \begin{minipage}{0.325\linewidth}
    \centering
    \includegraphics[width=\columnwidth,frame]{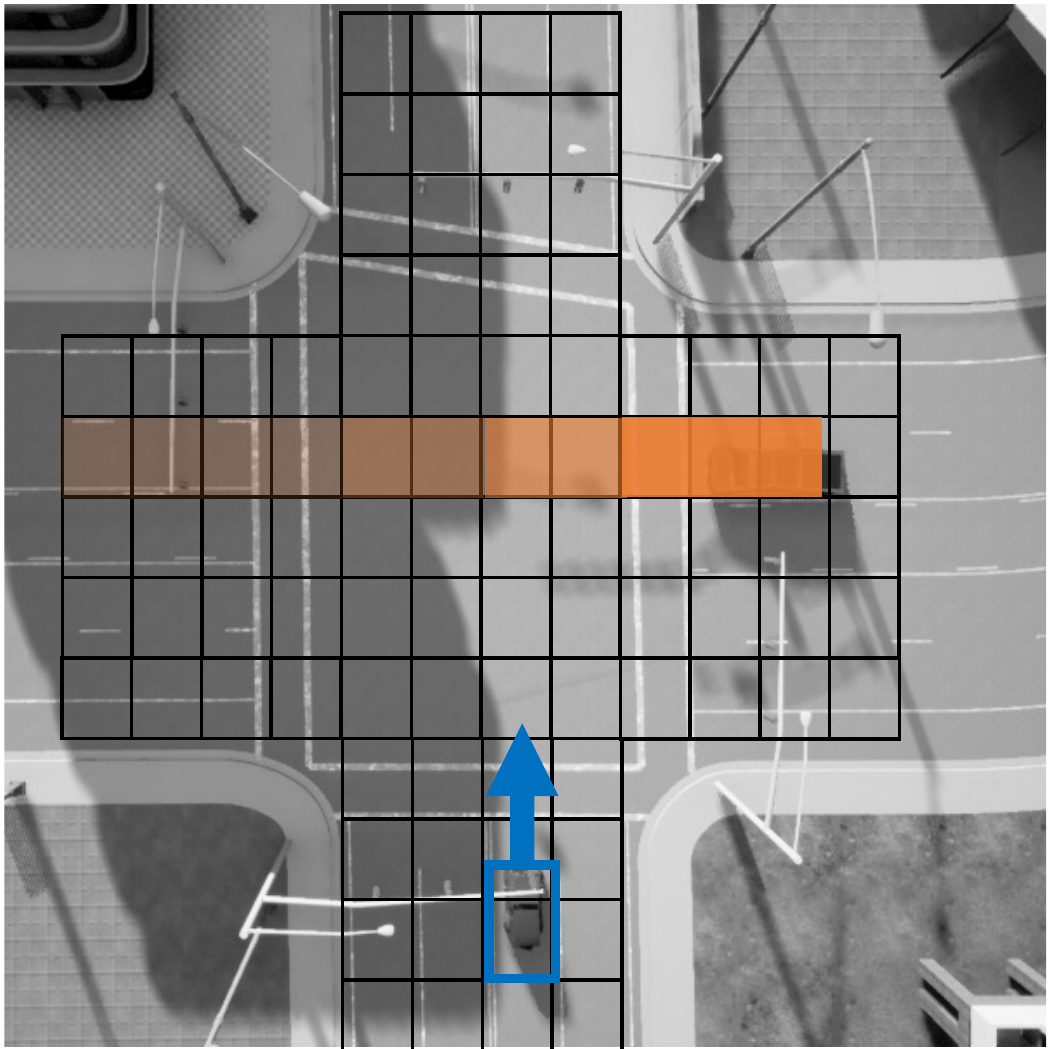}
    \\ {\small (b)}
  \end{minipage}
  \begin{minipage}{0.325\linewidth}
    \centering
    \includegraphics[width=\columnwidth,frame]{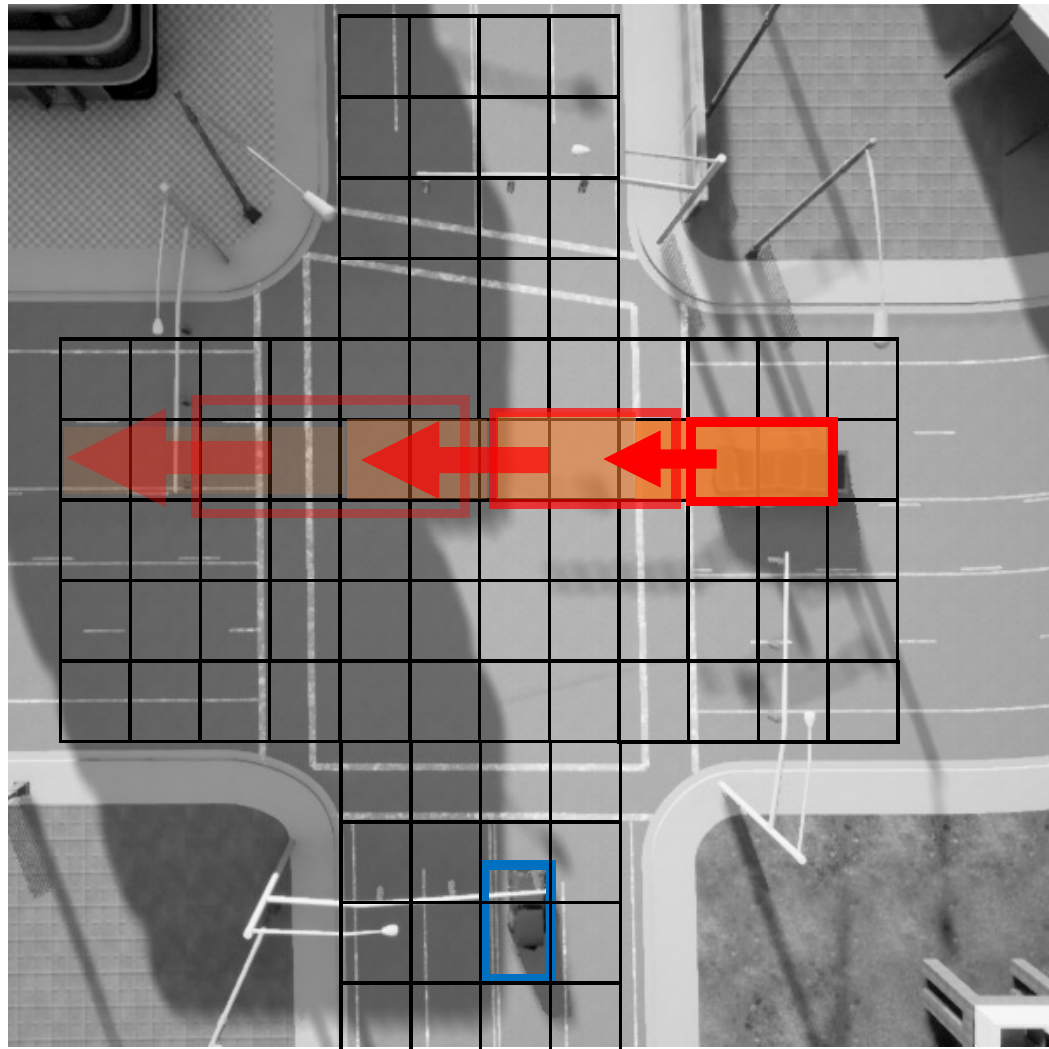}
    {\small (c)}
  \end{minipage}
  \caption{(a) The first planned vehicle (red) can use the entire drivable space. (b) The second vehicle (blue) treats the first as an obstacle (orange). Different shades represent different times at which grids are occupied. (c) The motion-adaptive buffer around a vehicle is proportional to its speed.}
  \label{fig:planning_examples}
\end{figure}

To demonstrate this, we have adapted a fast \textit{single-robot} motion planner, SIPP~\cite{SIPP}.
The input to SIPP is a goal for a robot, the positions over time of the dynamic obstacles and an occupancy grid of the environment (\cref{fig:planning_examples}(a)).
The output is a provably collision-free shortest path for the robot. 
At each frame, the planner must plan trajectories for every \sysname-capable vehicle.
Without loss of generality, assume that vehicles are sorted in some order.
The edge-offloaded planner iteratively plans trajectories for vehicles in that order: when running SIPP on the $i$-th vehicle, it represents all $i-1$ previously planned vehicles as dynamic obstacles in SIPP.
\cref{fig:planning_examples}(b) illustrates this, in which the trajectory of a previously planned vehicle is represented as an obstacle when planning a trajectory for the blue vehicle.

\begin{figure}[t]
  \centering
  \includegraphics[width=0.95\columnwidth,frame]{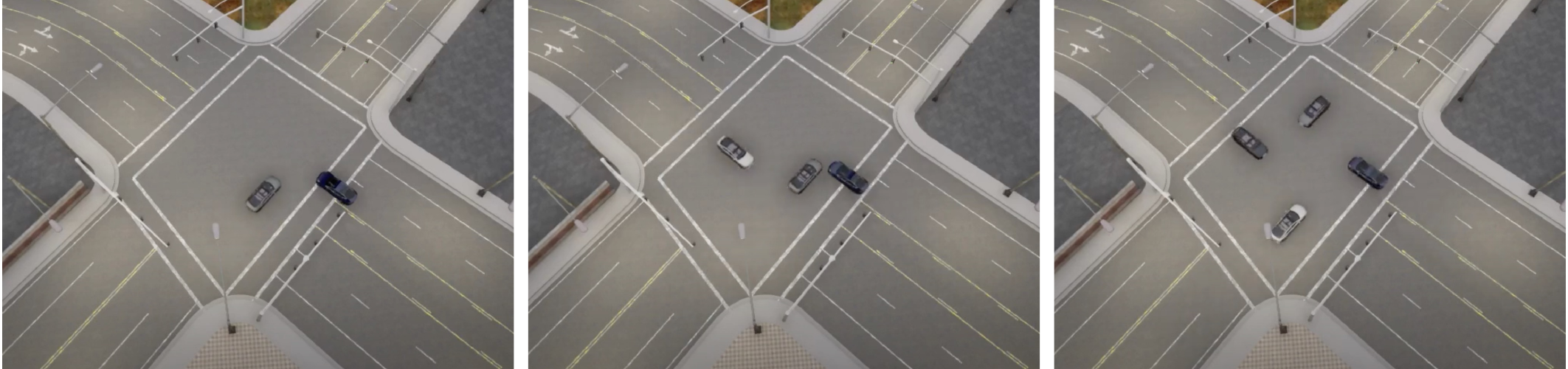}
  \caption{With autonomous driving's decentralized planning, in the absence of a traffic light controller, vehicles come to a deadlock at the intersection where they are unable to cross it.}
  \label{fig:deadlock}
\end{figure}

This approach has two important properties:
\begin{enumerate}[nosep,wide]

\item All trajectories are collision-free.
Two cars $i$ and $j$ cannot collide, since, if $j>i$, $j$'s trajectory would have used $i$'s as an obstacle, and SIPP generates a collision-free trajectory for each vehicle.

\item This planner cannot result in a traffic deadlock.
A deadlock occurs when there is a cycle of cars in which each car's forward progress is hampered by another.
\cref{fig:deadlock} shows examples of deadlocked traffic with two, three, and four cars (these scenarios under which these deadlocks occurred are described in \cref{s:high-thro-traff}).
However, our edge-offloaded planner cannot deadlock.
If there exists a cycle, there must be at least one pair of cars $i, j$ where $i < j$ and $i$ blocks $j$.
But this is not possible, because, when planning for $j$, the planner represents $i$'s trajectory as an obstacle.
\end{enumerate}

Because it was designed for robots, SIPP makes some idealized assumptions: all robots use the planned trajectories, all have the same dimensions, no trajectories are lost and robots can start and stop instantaneously.
In our implementation, we adapted SIPP to relax these but omit details for brevity.
\section{Evaluation}
\label{sec:evaluation}


Our evaluations demonstrate \sysname's performance and accuracy, and its potential for improving traffic safety and throughput.

\subsection{Methodology}
\label{sec:eval-methodology}

\begin{figure*}[t]
  \begin{minipage}{0.32\linewidth}
    \centering
     \includegraphics[width=0.74\columnwidth,frame]{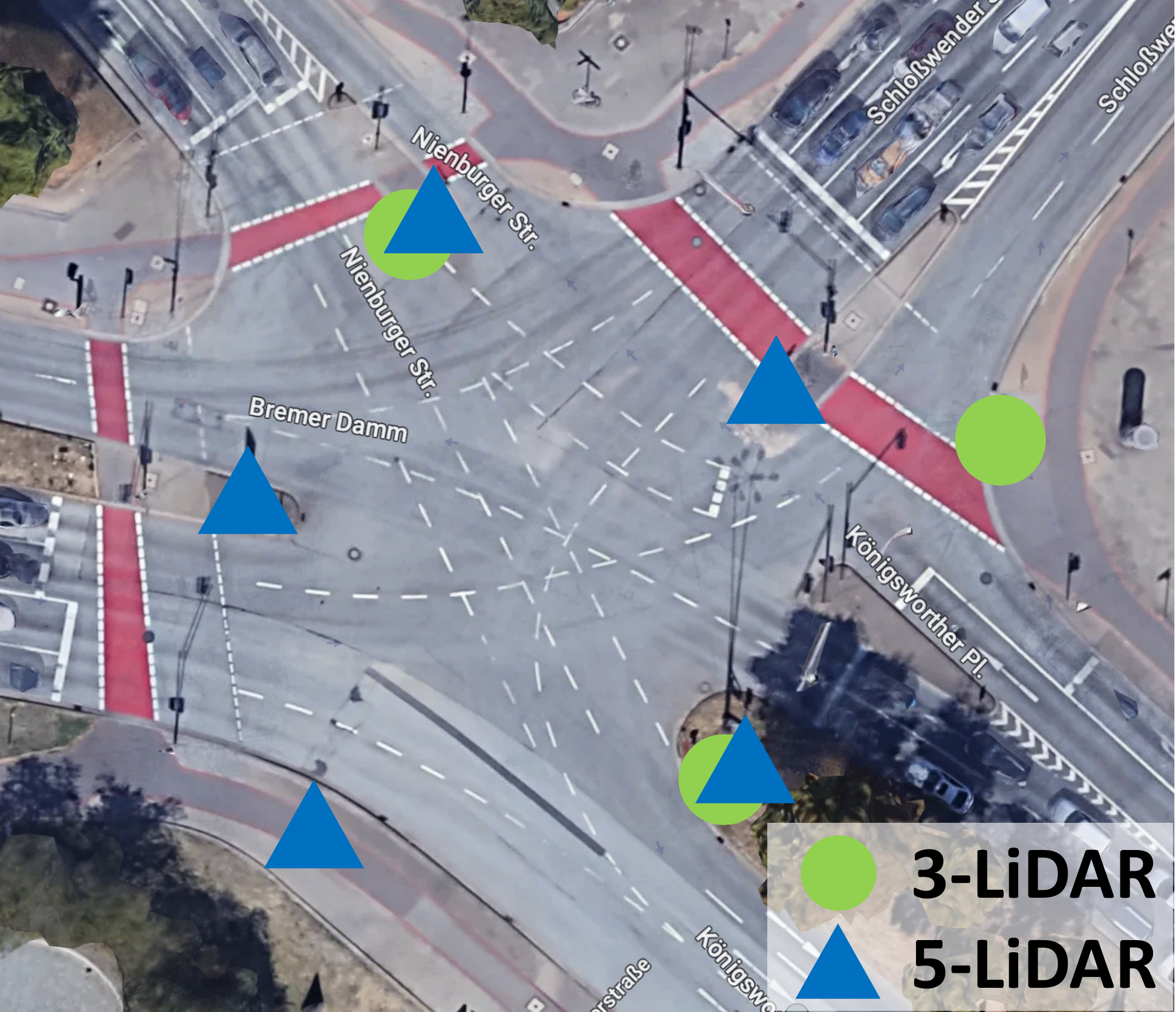}
       \caption{LiDARs placement in the LUMPI dataset.}
       \label{fig:lumpi_setup}
    \end{minipage}
  \begin{minipage}{0.33\linewidth}
      \centering
   \includegraphics[width=0.88\columnwidth]{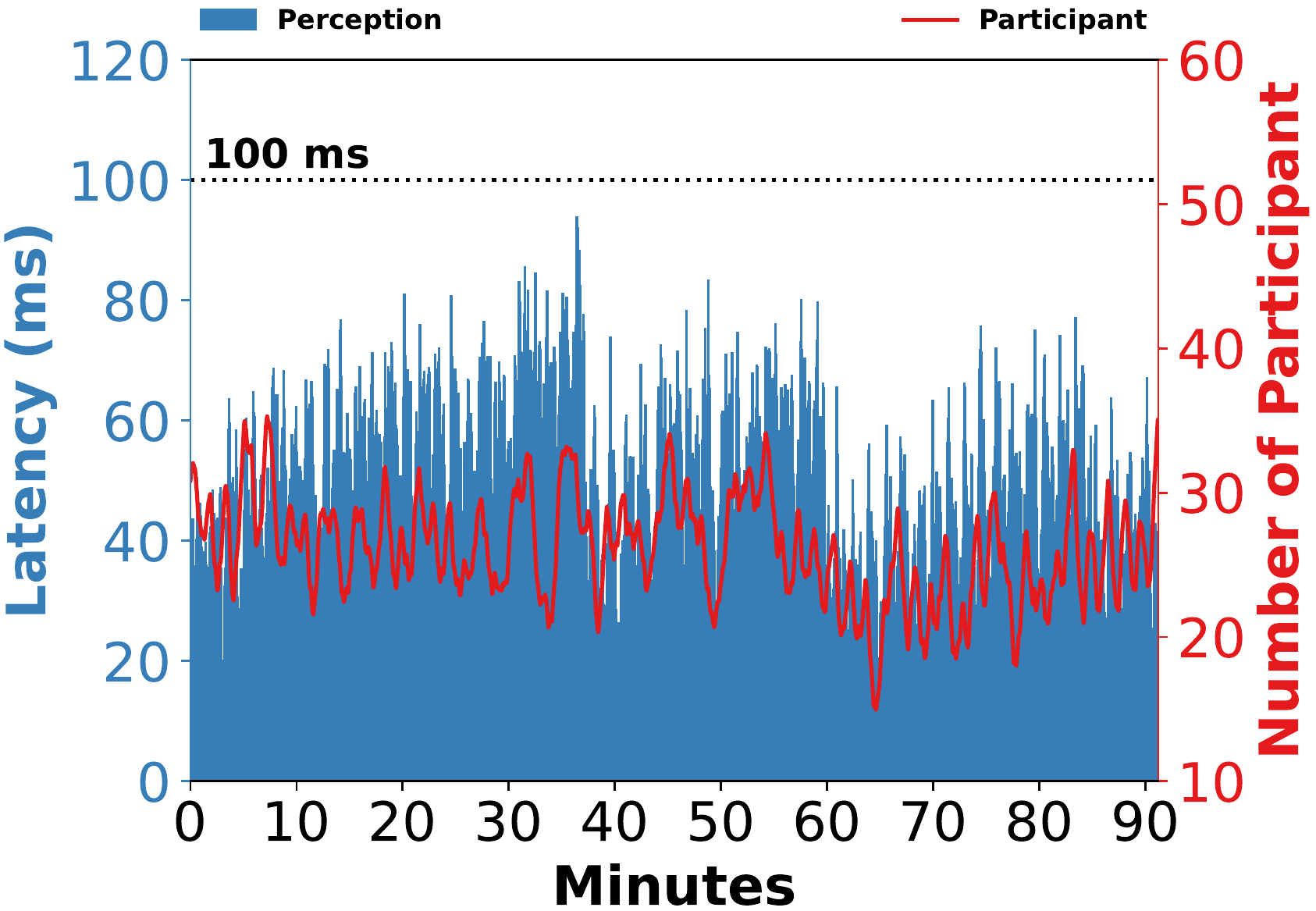}
      \caption{\sysname performance with a 3-LiDAR setup in the LUMPI dataset.}
    \label{fig:lumpi_3_lidars}
  \end{minipage}
  \begin{minipage}{0.33\linewidth}
  \centering
   \includegraphics[width=0.88\columnwidth]{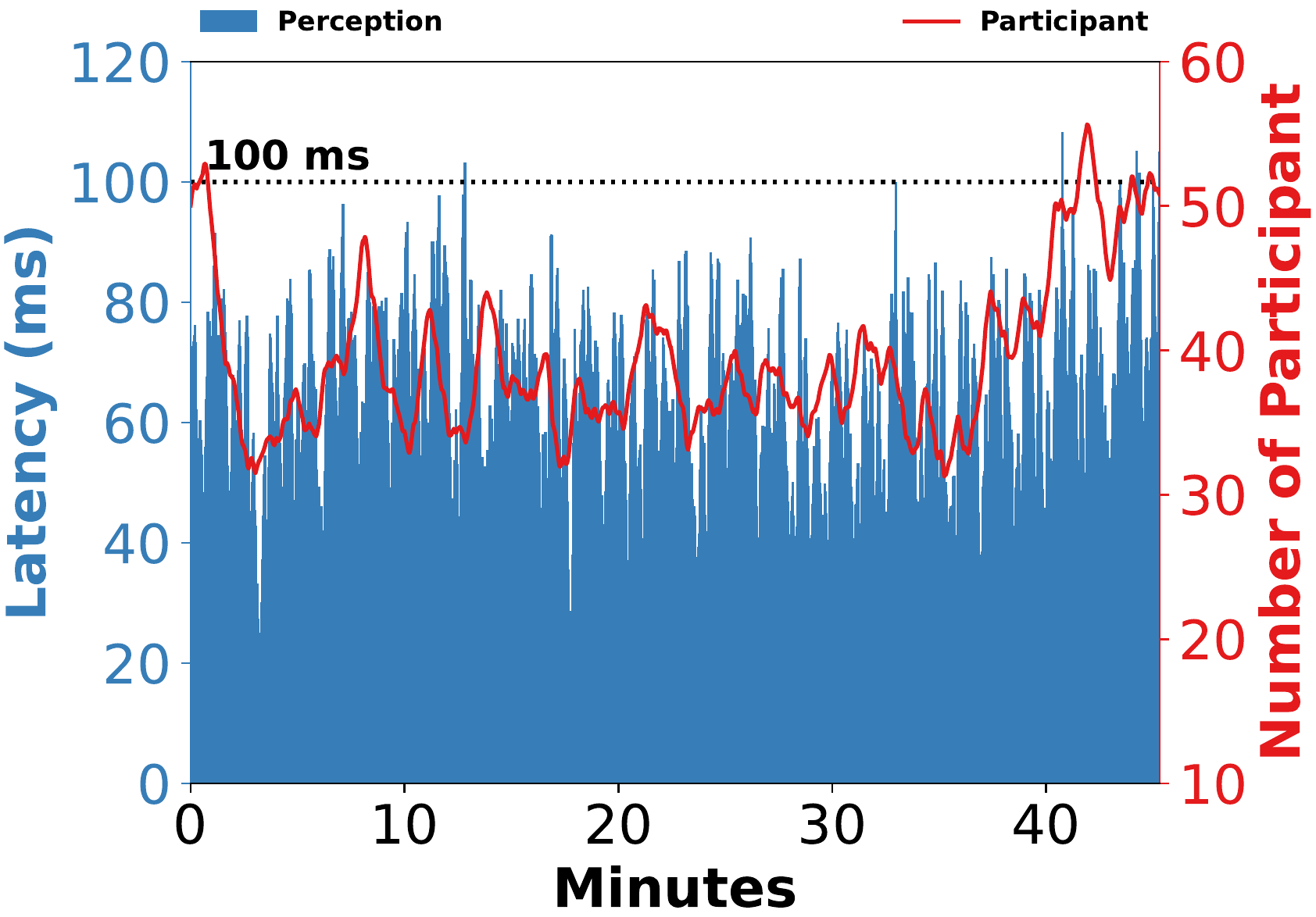}
     \caption{\sysname performance with a 5-LiDAR setup in the LUMPI dataset.}
     \label{fig:lumpi_5_lidars}
  \end{minipage}
\end{figure*}

\parab{Implementation.}
We implemented \sysname and the two use cases discussed in \cref{s:use_cases} on the Robot Operating System (ROS~\cite{quigley2009ros}).
ROS provides inter-node (ROS modules are called \textit{nodes}) communication using publish-subscribe, and natively supports points clouds and other data types used in perception-based systems.
\sysname runs as a ROS node that subscribes to point clouds, processes them as described in \cref{sec:design}, and publishes the results.
The offloaded planner (\cref{s:centralized_planning}) builds on top of an open-source SIPP implementation~\cite{sipp_impl}, runs as a ROS node, subscribes to the \sysname results, and publishes trajectories for each vehicle.
\sysname requires 6909 lines of C++ code, and the use cases 3800.

\parab{Real-world Traces.}
We evaluated \sysname on a large open source multi-LiDAR intersection dataset (LUMPI~\cite{busch2022lumpi}).
This dataset contains approximately 2.5~hours of 3D data collected across several days from upto five LiDARs mounted at a busy intersection in Hanover, Germany.
These LiDARs include two 16-beam LiDARs (Velodnye VLP-16) and three 64-beam LiDARs (Velodyne HDL-64, Hesai Pandar64, and Hesai PandarQT).
\sysname is designed to handle such sensor heterogeneity.

For all evaluations in this paper, we ran \sysname on an ``edge compute'' device, an AMD 5950x CPU (16 cores, 3.4 GHz) and a GeForce RTX 3080 GPU.
This device has significantly less compute than a commercial edge offering;
for example, a server on Google's distributed cloud edge has 96 vCPUs and 4 GPUs~\cite{GDCE}.
In that sense, our evaluation is conservative relative to what one might expect from a real deployment.

\parab{Real-world Testbed.}
Besides evaluating \sysname on real-world traces, we also built our own real-world testbed consisting of four Ouster LiDARs~\cite{Ouster} (an OS-1 64, an OS-0 128, and two OS-0 64).
Three of these have a field of view of $90^\circ$ while the last one has a $45^\circ$ field of view.
%

%

\parab{Simulation.}
The LUMPI dataset does not contain ground-truth, so we complement our evaluations with a simulator, which helps us evaluate \sysname accuracy and scaling.
We use CarLA~\cite{carlachallenge}, an industry-standard photo-realistic simulator for autonomous driving perception and planning.
It contains descriptions of virtual urban and suburban streets, and, using a game engine, can (a) simulate the control of vehicles in these virtual worlds, and (b) produce LiDAR point clouds of time-varying scenes.
Unless otherwise noted, our simulation based evaluations focus on intersections; several challenge scenarios for autonomous driving focus on intersections~\cite{nhsta2013}.

\parab{Metrics.}
We quantify end-to-end performance in terms of the 99th percentile of the latency (\textbf{\textit{p99 latency}}) between when a LiDAR generates a frame and when \sysname produces its outputs for that frame.
To quantify accuracy of individual components, we use metrics described in prior works (defined later).

%
%

\subsection{\sysname Performance}
\label{sec:eval-perf}

We ran \sysname on 2.25~hours of point clouds traces from the LUMPI dataset.
In these experiments, we measured the average, 99th percentile and end-to-end latency of \sysname.
Our results show that \sysname is able to achieve the 99th percentile latency less than 100~ms.

The LUMPI dataset provides two traces (\figref{fig:lumpi_setup}), one with three infrastructure LiDARs and another with five.
In the 3-LiDAR setup, the LiDARs are deployed on one side of a 4-way intersection.
The 5-LiDAR setup adds two more lidars to cover the other side of the intersection.
The latter covers more of the intersection, so \sysname detects more traffic participants.

\figref{fig:lumpi_3_lidars} and \figref{fig:lumpi_5_lidars} show the perception latency and number of participants for each frame.
In the 3-LiDAR setup, \sysname's median latency is 38.7ms and p99th latency is 71.6ms.
In the 5-LiDAR setup, \sysname's median latency is 56.1ms and p99th is 88.8ms.
\sysname's latency roughly scale with the number of participants in the intersection.
In the trace with the 5-LiDAR setup, a small number of frames exceed the 100ms target; these frames have more than 50 participants and more than 30 vehicles.
The number of participants in the 5-LiDAR setup is about 1.5$\times$ more than the 3-LiDAR setup.
Overall, it shows that \sysname can comfortably support, with modest computing resources, such an busy intersection with more than 40 participants and can still maintain the tail latency within 100~ms.


\subsection{\sysname Accuracy}
\label{sec:eval-microbenchmark}

Because the LUMPI dataset 
does not have ground-truth, we evaluated \sysname's accuracy using real-world data that we collected using our own testbed.\footnote{We also use this dataset to quantify latency with offloaded planning; \cref{s:simul-results:-laten}.}
On this dataset, we manually labeled ground-truth positions.
We also evaluated \sysname's accuracy using CarLA.
We show that \sysname's positioning, heading, speed, and tracking accuracies are comparable to that reported in other works.

\parab{Metrics.}
We report positioning error (in m), which chiefly depends upon the accuracy of alignment.
We measure heading accuracy using the average deviation of the heading vector, in degrees, from ground truth.
For the accuracy of velocity estimation, we report both absolute and relative errors.
Finally, we use two measures to capture tracking performance~\cite{mot_metric}: multi-object tracking accuracy (MOTA) and precision (MOTP).
The former measures false positives and negatives as well as ID switches (\cref{sec:design}); the latter measures average distance error from the ground-truth track.

\parab{Results.}
\cref{tab:accuracy_results} summarizes our findings.
%
Positioning error is about 8-10~cm in \sysname, both in simulation and in the real world; the state-of-the-art LiDAR SLAM~\cite{vloam} reports about 15~cm error.
Our heading estimates are comparable to prior work that uses a neural network to estimate heading.
Speed estimates are highly accurate, both in an absolute sense (error of a few cm/s) and in a relative sense (over 97\%).


Finally, tracking is also highly accurate.
In the real-world experiment, tracking was perfect.
In simulation, with 10 vehicles concurrently visible, MOTA is over 99\%, and \sysname outperforms the state-of-the-art neural network in 3D tracking~\cite{mono3dt}, for two reasons.
The neural network solves a harder problem, tracking from a moving LiDAR.
Our fused LiDAR views increase tracking accuracy; when using a single LiDAR to track, MOTA falls to 93\%.
Finally, MOTP is largely a function of positioning error, so it is comparable to that value.

\parab{Alignment Accuracy.}
Although alignment is performed only once, its accuracy is crucial for \sysname; without accurate alignment, \sysname's perception components could not have matched the state-of-the-art (\cref{tab:accuracy_results}).

\begin{table}[t]
\centering
\begin{footnotesize}
\begin{tabular}{c|c|c|c} \hline
  \textbf{Metric} & \textbf{Real} & \textbf{Sim} & \textbf{Prior} \\ \hline \hline
  Positioning Error (m) & 0.10 & 0.08 & 0.15~\cite{vloam, cvivsic2022soft2}   \\ \hline
  Heading Error ($^\circ$) & 8.17 & 6.45 & 5.10~\cite{intentNet}   \\ \hline
  Speed Error (m/s) & 0.04 & 0.06 & --- \\ \hline
  Speed Accuracy (\%) & 98.80 & 97.49 &  ---  \\ \hline
  MOTA (\%) & 100 & 99.54 & 84.52~\cite{mono3dt}    \\ \hline
  MOTP (m) & 0.12 & 0.08 & ---  \\ \hline
\end{tabular}
\end{footnotesize}
\caption{Perception accuracies from real-world data and simulation compared to prior works.}
\label{tab:accuracy_results}
\end{table}

\parae{Comparison alternatives.}
To contextualize \sysname's alignment performance, we compared it against two other ways of obtaining an initial guess for ICP: a standard feature-based approach, SAC-IA~~\cite{rusu09:_fast_point_featur_histog_fpfh}; and using GPS. 
In this experiment, we use point clouds from our simulations and real-world experiments.
The three approaches estimate initial guesses for the pose of three LiDARs, and feed those poses to ICP for building a stitched point cloud.
In our evaluations, we report the average root mean square error (RMSE) between the stitched point cloud (\textit{after} running ICP on the initial guesses) for every approach against a ground-truth.

\parae{Results.}
\sysname's alignment results in errors of a few cm (\cref{tab:alignment_accuracy}), almost 2-3 orders of magnitude lower error than the competing approaches, which explains why we chose this approach.
SAC-IA~\cite{rusu09:_fast_point_featur_histog_fpfh} does not take any inputs other than the point clouds and estimates the transformation between two point clouds using 3D feature matching.
However, this works well only when point clouds have a large amount of overlap.
In our setting, LiDARs are deployed relatively far from each other resulting in less overlap, and SAC-IA is unable to extract matching features from multiple LiDAR point clouds.
Using GPS for alignment provides a good initial guess for the relative \textit{translation} between the LiDARs.
However, GPS cannot estimate the relative \textit{rotation} between LiDARs, so results in poor accuracy.
 
\begin{table}[t]
\centering
\begin{tabular}{cc|cc}
\hline
\multicolumn{2}{c|}{\multirow{2}{*}{}}                           & \multicolumn{2}{c}{\textbf{RMSE (m)}}                            \\ \cline{3-4} 
\multicolumn{2}{c|}{}                                            & \multicolumn{1}{c|}{Average}            & Std Dev       \\ \hline\hline
\multicolumn{1}{c|}{\multirow{3}{*}{Simulation}} & \textbf{\sysname} & \multicolumn{1}{c|}{\textbf{0.03}} & \textbf{0.02} \\ \cline{2-4} 
\multicolumn{1}{c|}{}                            & SAC-IA        & \multicolumn{1}{c|}{39.2}         & 13.4         \\ \cline{2-4} 
\multicolumn{1}{c|}{}                            & GPS           & \multicolumn{1}{c|}{23.8}         & 11.1         \\ \hline
\multicolumn{1}{c|}{\multirow{3}{*}{Real-world}}  & \textbf{\sysname} & \multicolumn{1}{c|}{\textbf{0.09}} & \textbf{0.04} \\ \cline{2-4} 
\multicolumn{1}{c|}{}                            & SAC-IA        & \multicolumn{1}{c|}{11.8}         & 13.9         \\ \cline{2-4} 
\multicolumn{1}{c|}{}                            & GPS           & \multicolumn{1}{c|}{13.0}         & 1.2          \\ \hline
\end{tabular}
\caption{\sysname's novel alignment algorithm outperforms existing state-of-the-art initial alignment algorithms.} 
\label{tab:alignment_accuracy}
\end{table}




\subsection{Latency Breakdown and Scaling} 
\label{s:simul-results:-laten}

\parab{Setup.}
To explore the total latency with more vehicles and to understand the breakdown of latency by component, we designed several scenarios in CarLA with increasing numbers of vehicles traversing a 4-way intersection.
%
%
At this intersection, both streets have two lanes in each direction.
We varied the number of vehicles from 2 to 14, to understand how \sysname's components scale.
To justify this range of the number of vehicles, we use the following data: the average car length is 15~ft~\cite{car_length} and the width of a lane is 12~ft~\cite{lane_width}.
Allowing for lane markers, medians, and sidewalks, let us conservatively assume that the intersection is 60~ft across.
Suppose the intersection has traffic lights.
Then, if traffic is completely stalled or moving very slowly, at most 4 cars can be inside the intersection per lane, resulting in a maximum of 16 cars (\ie the maximum capacity of the intersection is 16).
If cars are stalled, \sysname does not incur much latency since it does not have to estimate motion, heading, or plan for these, so we limit our simulations to 14 vehicles.\footnote{We have verified that, above 14 vehicles, end-to-end latencies actually drop.}
On the other hand, if traffic is moving at 45~mph (or 66~fps) and cars maintain a 3-second~\cite{three_second} safe following distance, then at most one car can be within the intersection per lane, for a total of 4 cars.


\begin{table}[t]
\footnotesize
\centering
\begin{tabular}{c|c|c|c|c|c|c}
\hline
 & \multicolumn{6}{c}{\textbf{Number of Vehicles}} \\ \hline \hline
\textbf{Component} & \textbf{2} & \textbf{4} & \textbf{7} & \textbf{10} & \textbf{12} & \textbf{14} \\ \hline
\begin{tabular}[l]{@{}l@{}}BG Subtraction\end{tabular} & 7.5 & 8.9 & 9.9 & 9.9 & 11.0 & 18.7 \\ \hline
Stitching & 0.08 & 0.11 & 0.13 & 0.13 & 0.17 & 0.19 \\ \hline
Clustering & 4.3 & 5.8 & 11.2 & 13.0 & 22.2 & 20.5 \\ \hline
  \begin{tabular}[l]{@{}l@{}}Bounding
    Box\end{tabular} & 0.06 & 0.07 & 0.09 & 0.15 & 0.17 & 0.16 \\ \hline
Tracking & 0.1 & 0.15 & 0.21 & 0.28 & 0.37 & 0.38 \\ \hline
\begin{tabular}[l]{@{}l@{}}Heading Vector\end{tabular} & 8.8 & 12.9 & 24.0 & 28.8 & 41.9 & 52.6 \\ \hline
\textbf{Total} & \textbf{18.5} & \textbf{26.4} & \textbf{43.7} & \textbf{47.1} & \textbf{69.9} & \textbf{81.5} \\ \hline
\end{tabular}
\caption{p99 per-frame latency (in milliseconds) for perception. We exclude latency numbers for motion vector estimation which are on the order of a few microseconds.}
\label{tab:latency-ours}
\end{table}

\parab{Breakdown for \sysname.}
\cref{tab:latency-ours} depicts the breakdown of 99-th percentile (p99) latency by component for \sysname as well as the total p99 latency, as a function of the number of vehicles in the scene.
In all our experiments, \sysname processed frames at the full frame rate (10~fps).

The total p99 latency for \sysname increases steadily up to 82~ms for 14 vehicles\footnote{For many of our experiments, including this one, we have generated videos to complement our textual descriptions.
These are available at an \textit{\textbf{anonymous YouTube channel}}: \url{https://www.youtube.com/@cip-iotdi24}.} from 19~ms for 2 vehicles.
This highlights perception's data dependency (\cref{sec:design}); performance of some components depends on the number of participants.
%
%
These numbers suggest that modest off-the-shelf compute hardware that we have used in our experiments might be sufficient for traffic management at moderately busy intersections.
This data dependency also suggests that deployments of \sysname will need to carefully provision their infrastructures based on historical traffic (similar to network planning and provisioning).

The three most expensive components are background subtraction, clustering and heading vector estimation.
Background subtraction accounts for about 10~ms, but depends slightly on the number of vehicles; to be robust, it uses a filter (details omitted in \cref{s:fusion}) that is sensitive to the number of points (or vehicles).
Clustering accounts for about 20~ms with 14 vehicles and is strongly dependent on the number of vehicles since each vehicle corresponds to a cluster.

Heading estimation accounts for nearly 65\% of perception latency, even after GPU acceleration (\cref{s:tracking}).
These results show that heading vector estimation not only depends on the number of vehicles, but on their dynamics as well.
When we ran perception on 16 vehicles, p99 latency actually dropped; in this setting, 16 vehicles congested the intersection, so each vehicle moved very slowly.
Heading vector estimation uses ICP between successive object point clouds; if a vehicle hasn't moved much, ICP converges faster, accounting for the drop.

Other components are lightweight.
Stitching is fast because of the optimization described in \cref{s:fusion}.
Bounding box estimation is inherently fast.
Track association is cheap because it tracks a single point per vehicle, the centroid of the bounding box.
Motion estimation takes a few microseconds and relies on positions computed during stitching.
Thus, leveraging abstractions and reusing values from earlier in the pipeline help \sysname meet latency targets (\cref{s:tracking}).

\parae{Benefits of optimizations.}
\cref{tab:unopt_opt} quantifies the benefits of our optimizations.
Stitching before background subtraction requires nearly 70~ms in total; reversing the order reduces this time by 6.7$\times$.
By exploiting LiDAR characteristics (\cref{s:fusion}), \sysname can perform background subtraction in 1.5~ms per frame.\footnote{\cref{tab:latency-ours} does not include this optimization, since it can only be applied to some LiDARs}
A CPU-based heading vector estimation requires nearly 1~s which would have rendered \sysname infeasible; GPU acceleration (\cref{s:tracking}) reduces latency by 35$\times$.

\parab{Calibration Steps}.
Finally, alignment (\cref{s:fusion}) of 4 LiDARs takes about 4 minutes.
This includes not just the time to guess initial positions, but to run the ICP (on a CPU).
Because it is invoked infrequently, we have not optimized it.

\begin{table}[t]
\centering
\footnotesize
\begin{tabular}{c|c|c|c}
\hline
\multicolumn{1}{c|}{\textbf{Optimization}} & \textbf{Before} & \textbf{After} & \textbf{Ratio} \\ \hline \hline
  BG subt before stitching & 67.7 & 10.0 & 6.7 \\ \hline
  Exploiting LiDAR characteristics & 9.9 & 1.5 & 6.6 \\ \hline
  Heading vector GPU acceleration  & 1057.7 & 28.8 & 36.6 \\ \hline
\end{tabular}
\caption{Impact of optimizations on p99 latency}
\label{tab:unopt_opt}
\end{table}

\subsection{Perception Augmentation: Safety}
\label{s:bett-traff-manag}

In this and subsequent sections, we quantify the feasibility and benefits of the use cases in \cref{s:use_cases}.
We begin by demonstrating the increased safety resulting from augmenting an autonomous vehicle's perception with \sysname outputs (\cref{s:augment_vehicle_perception}).
\sysname has a comprehensive view of an intersection, so it can lead to increased safety.
To demonstrate this, we implemented two scenarios in CarLA from the US National Highway Transportation Safety Administration (NHTSA) precrash typology~\cite{nhsta2013}; these are challenging scenarios for autonomous driving~\cite{carlachallenge}.

\parae{Red-light violation.}
A orange truck and the ego-vehicle (yellow box) approach an intersection (\cref{fig:red_light_violation}).
An oncoming vehicle (red box) on the other road violates the red traffic light.
The orange truck can see the violator and hence avoid collision, but the ego-vehicle cannot.

\parae{Unprotected left-turn.}
The ego-vehicle (yellow box) heads towards the intersection (\cref{fig:unprotected_left_turn}).
A vehicle (red box) on the opposite side of the intersection makes an unprotected left-turn.
The ego-vehicle's view is blocked by the orange trucks.

\parab{Methodology and Metrics.}
In each scenario, \sysname augments the ego-vehicle's on-board perception.
When comparing against (un-augmented) autonomous driving, to ensure a more-than-fair comparison, we (a) equip autonomous driving with ground-truth (perfect) perception and (b) use an on-board SIPP planner in single-mode (plan only for the ego-vehicle) for autonomous driving.
The alternative would have been to use an open-source stacks like Autoware~\cite{Autoware} which has its own perception and planning modules.
However, in these experiments, we are trying to understand the impact of augmenting a vehicle's perception with \sysname, so we chose a simpler approach that equalizes implementations.

For both scenarios, we vary speeds and positions of the ego-vehicle and oncoming vehicle to generate 16 different experiments.
We then compare for what fraction of experiments each approach can guarantee safe passage.

\parab{Results.}
In both scenarios (\cref{tab:end_to_end_experiments}), autonomous driving ensures safe passage in fewer than 20-40\% of the cases.
\sysname ensures safety in all cases because it senses the oncoming traffic that is occluded from the vehicle’s on-board sensors\footnote{\label{youtube}Please see YouTube channel for videos. \url{https://www.youtube.com/@cip-iotdi24}}.
This gives the planner enough time to react and plan a collision avoidance maneuver --- in this case, stop the vehicle.
Of the two cases, the unprotected left-turn was the more difficult one for \sysname as it is for autonomous driving, which fails more often in this case.
Yet, \sysname is able to guarantee safe passage in all 16 cases.
In the red-light violation scenario, \sysname senses the oncoming traffic early on and has enough time to react.
However, in the unprotected left-turn, the ego-vehicle is traveling relatively fast and the oncoming traffic takes the left-turn at the last moment.
Even in this case, \sysname gives the vehicle enough time to react.
Though \sysname has a smaller time to react, its motion-adaptive bounding box and stopping distance estimation ensure that the vehicle stops on time.

\begin{table}[t]
\centering
\footnotesize
\begin{tabular}{c|c|c}
\hline
\multirow{2}{*}{\begin{tabular}[c]{@{}c@{}}\textbf{NHTSA}\\ \textbf{Scenario}\end{tabular}} & \multicolumn{2}{c}{\textbf{Safe Passage (\%)}} \\ \cline{2-3} 
                      & \textit{\sysname} & \textit{Autonomous Driving} \\ \hline \hline
Red-light Violation   & 100                     & 37                 \\ \hline
Unprotected Left-turn & 100                     & 18                 \\ \hline
\end{tabular}
\caption{With more comprehensive perception, \sysname can provide safe passage to vehicles in both scenarios.}
\label{tab:end_to_end_experiments}
\vspace{-2mm}
\end{table}





\begin{figure}[t]
\begin{minipage}{0.49\linewidth}
    \centering
 \includegraphics[width=0.99\columnwidth]{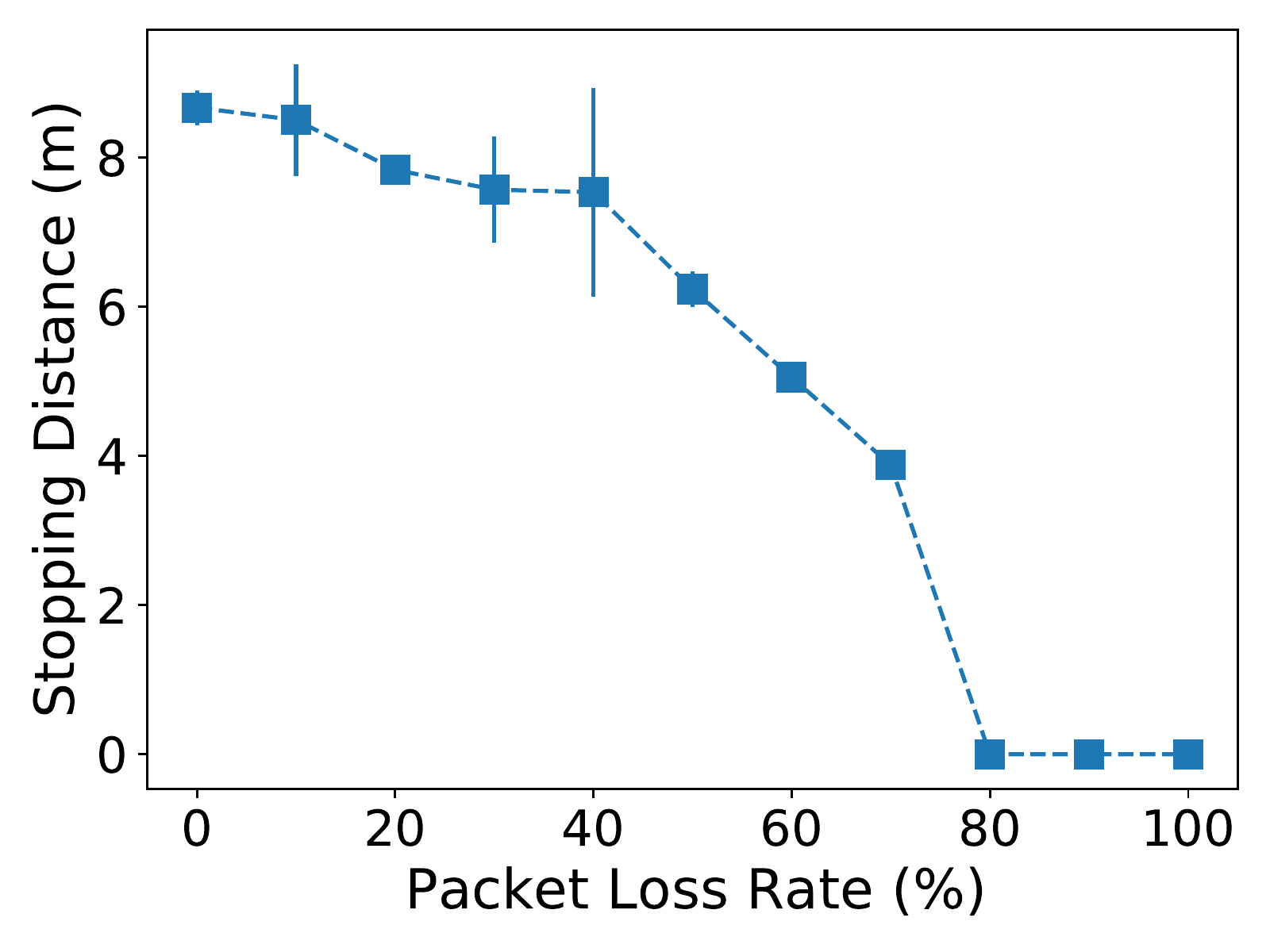}
    \caption{\sysname can ensure collision-free trajectories with up to 70\% packet-loss rates.}
  \label{fig:packet_loss}
\end{minipage}
\begin{minipage}{0.49\linewidth}
\centering
 \includegraphics[width=0.99\columnwidth]{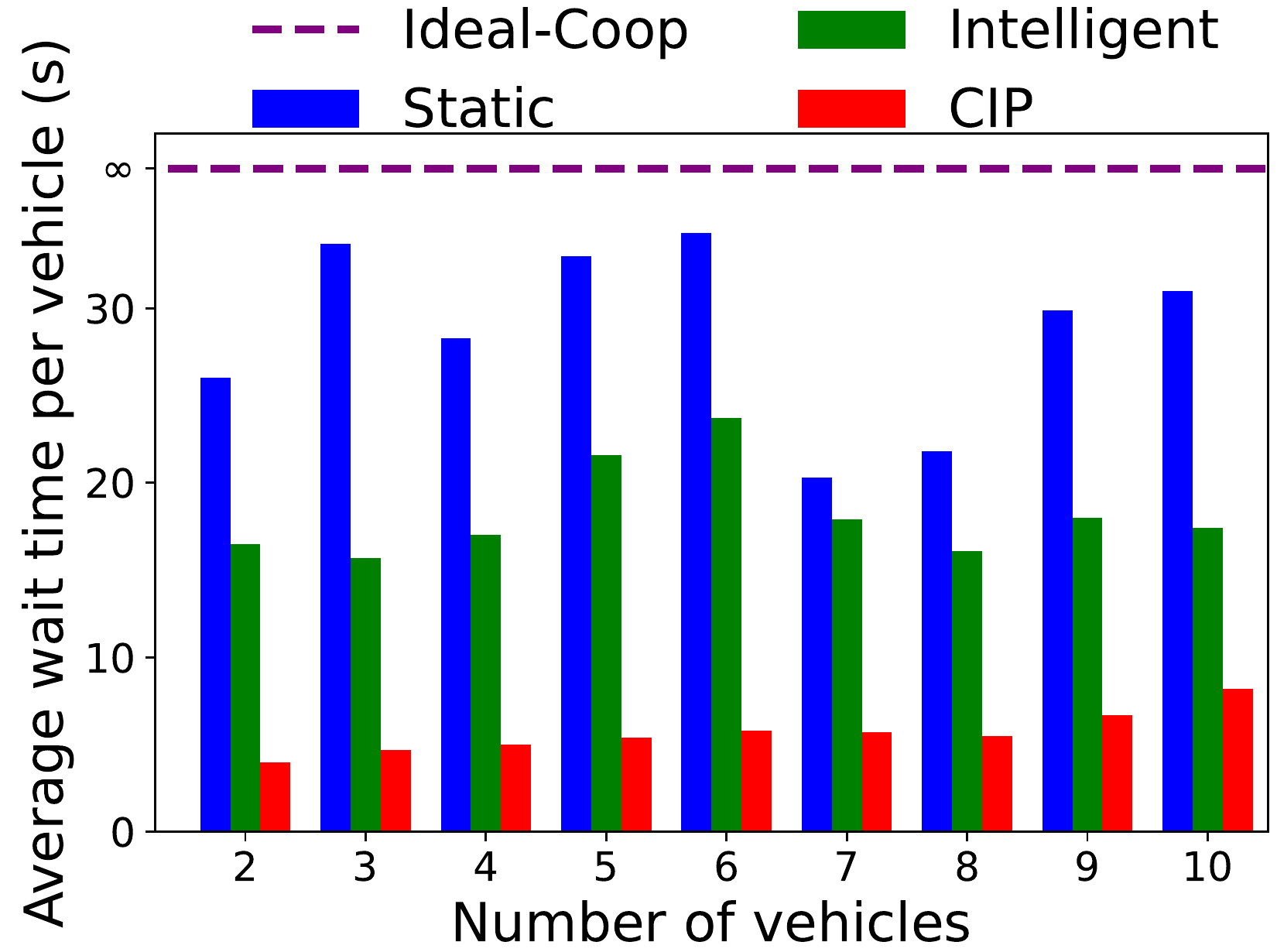}
   \caption{\sysname minimizes average wait times (seconds) for vehicles with traffic light free intersections.}
   \label{fig:traffic_light}
\end{minipage}
\end{figure}

\parab{Robustness to Packet Loss.}
In our implementation, the centralized planner transmits trajectories wirelessly to vehicles. It generates trajectories over a longer time horizon (\cref{s:centralized_planning}) to be robust to packet loss. To quantify its robustness, we simulated packet losses ranging from 0 to 100\% for in the red-light violation scenario (\cref{fig:red_light_violation}). For each loss rate, we measure the stopping distance between the ego-vehicle and the oncoming traffic which violates the red-light. Higher stopping distances are good (\cref{fig:packet_loss}). 
As we increased the packet-loss, the stopping distance decreased because the ego-vehicle was operating on increasingly stale information. Even so, \sysname ensures collision-free passage for the ego-vehicle through the intersection till 70\% loss, with minimal degradation in stopping distance till about 40\% loss.

\subsection{Offloaded Planner: Latency}
\label{sec:eval-real_world_deployment}

In this section, we measure the performance of a real-world deployment with \sysname running an offloaded planner on the same device as the \sysname stack.
To do this, we deployed four LiDARs at the corners of a busy four-way intersection (\cref{fig:intersection_example}(a)) with heavy pedestrian and vehicular traffic in a large metropolitan area.
These LiDARs connected to the edge compute device using Ethernet cables.
The edge compute connected to Raspberry Pis on the vehicles through Wi-Fi (to proxy 5G).
We collected data for nearly 30 minutes; we measured and report the end-to-end latency for every frame.

\parab{Metrics.}
We measure the end-to-end latency of \sysname and the centralized planner.
This is the time from when \sysname receives 3D point clouds to when a vehicle receives its trajectory from the centralized planner over the wireless network.

\begin{figure}[t]
  \centering
  \includegraphics[width=0.73\columnwidth]{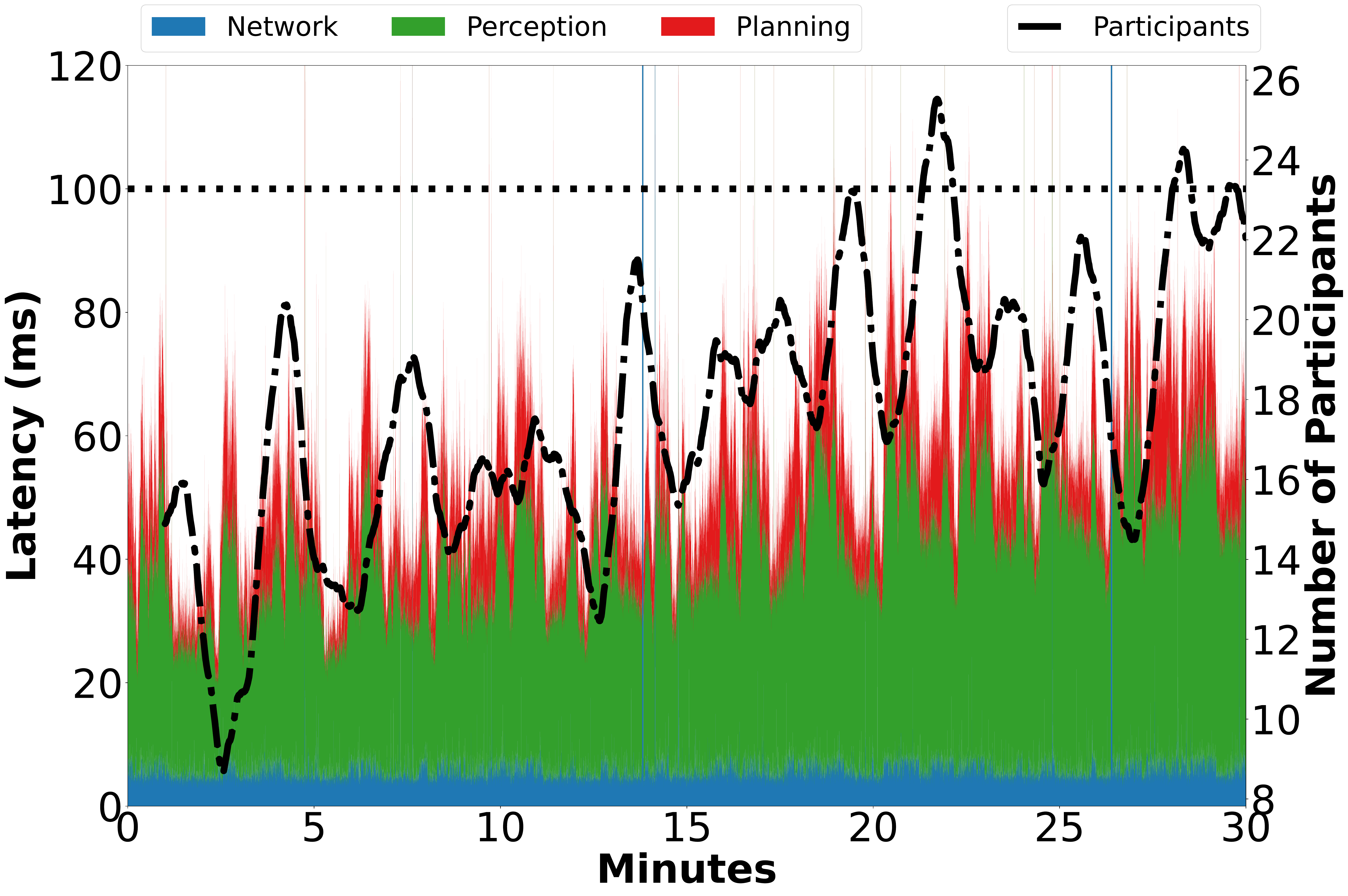}
  \caption{Per-component end-to-end latency (left y-axis) along with the number of traffic participants (right y-axis) from a deployment of \sysname at a busy intersection in the real-world for over 30 minutes.}
  \label{fig:end_to_end_latency}
\end{figure}

\parab{Results.}
\cref{fig:end_to_end_latency} shows the end-to-end latency for each frame, for over 30 minutes (approximately 18,000 frames), broken down by component.
The average end-to-end latency is 57~ms and p99 latency is 91~ms.
This shows that \sysname can operates under the 100~ms latency budget for  autonomous driving used by Mobileye\cite{mobileye}.
Moreover, \sysname processed LiDAR input at full frame rate.
Lastly, unlike autonomous driving pipelines today which plan for only a single vehicle, \sysname with offloaded perception can plan for 10~s of vehicles simultaneously within the 100~ms latency budget.

On average, \sysname detects 18 traffic participants 
per frame at the intersection during our experiment.
As our experiment progressed, the traffic at the intersection steadily increased, as shown by the dotted line in \cref{fig:end_to_end_latency} (representing the running average of traffic participants for 600~frames or 60~seconds).
Because \sysname's latency depends on the number of traffic participants, this contributed to the slow increase in end-to-end latency towards the end of the experiment.

From this graph, we also observe that network latency is small \ie the average is 9~ms, whereas p99 is 17~ms.
The same is true for planning latency for which the average is 11~ms and p99 is 26~ms.
Planning latency scales with the number of vehicles.
As the number of vehicles increase through the course of the experiment, planning latency also increases.
However, in overall, perception latency (37~ms in average and p99 of 60~ms) dominates, which motivates the careful algorithmic and implementation choices in \cref{sec:design}.


\parab{Scaling Offloaded Planning.}
To better understand how offloaded planning scales with the number of participants, \cref{tab:latency_planning} depicts the p99 planning latency from simulations.
As expected, there is a dependency on the number of vehicles, since \sysname individually plans for each vehicle.
Planning latencies can be slightly non-monotonic -- the planning cost for 12 vehicles is more than that for 14 -- because the planner's graph search can depend upon the actual trajectories of the vehicles, not just their numbers.




\subsection{Offloaded Planner: High-Throughput Traffic Management}
\label{s:high-thro-traff}

In this section, we demonstrate \sysname's use of offloaded planning to improving traffic throughput.
Intersections contribute significantly to traffic congestion~\cite{AIMSurvey}; \textit{\textbf{traffic-light free intersections}} can reduce congestion and wait times.
An offloaded planner, enabled by \sysname, because it centrally plans trajectories for all \sysname-capable vehicles, can plan collision-free trajectories at an intersection without traffic lights.
We have verified this in simulations as well.

\parab{Wait-time Comparison.}
A traffic-light free intersection can significantly reduce wait times at intersections, thereby enabling higher throughput.
To demonstrate this, we compared \sysname's average wait times (using a centralized planner) at an intersection against three other approaches:
\begin{itemize}[topsep=0pt,itemsep=-1ex,partopsep=1ex,parsep=1ex,leftmargin=*]
\item \textit{Static.} An intersection with static traffic lights.
\item \textit{Intelligent.} Intelligent traffic light control~\cite{intelligent_tl} which prioritizes longer queues.
\item \textit{Ideal-Coop.} A traffic-light free intersection where autonomous vehicles use centralized perception but on-board planning (\cref{s:augment_vehicle_perception}).
This represents an idealized version of cooperative perception~\cite{autocast} because it assumes that every vehicle can see  all other traffic participants.
\end{itemize}


For the first two approaches, we obtained policies from published best practices~\cite{natco}. In all experiments, we placed four LiDARs at an intersection in CarLA. 

\parab{Results.}
Compared to \textit{Static} and \textit{Intelligent}, \sysname reduces average wait times for vehicles by up to 5$\times$ (\cref{fig:traffic_light}).
It performs better than even \textit{Intelligent} because it can minimize the stop and start maneuvers at the intersection (by preemptively slowing down some vehicles), and hence can increase overall throughput, leading to lower wait times.\footnoteref{youtube}

Interestingly, beyond about 6 vehicles, wait times for strategies that use traffic lights (\textit{Static} and \textit{Intelligent}) drop.
Recall that our intersection has two lanes in each direction.
As the number of vehicles increases, the probability that a vehicle can pass the intersection without waiting increases.
For example, with \textit{Intelligent}, a vehicle waiting at an intersection can trigger a green light, so a vehicle arriving in the adjoining lane can freely pass.
Even so, \sysname has lower wait times than other alternatives up to 10 vehicles (wait time comparisons are similar beyond this number, results omitted for brevity).

\cref{fig:traffic_light} shows infinite wait times for \textit{Ideal-Coop}.
That is because, although one might expect that a decentralized planner might have comparable throughput to \sysname, we found that it \textbf{\textit{led to a deadlock}} (\cref{s:centralized_planning}) with as few as two vehicles (each vehicle waited indefinitely for the other to make progress, resulting in zero throughput).\footnoteref{youtube}
Fundamentally, this occurs because, with \textit{Ideal-Coop}, vehicles lack global knowledge of planning decisions.
Thus, deadlocks can happen at intersections without traffic lights even for more practical cooperative perception approaches which use on-board planners~\cite{autocast}.

\begin{table}[t]
\centering
\footnotesize
\begin{tabular}{c|c|c|c|c|c|c}
\hline
\textbf{Number of Vehicles} & 2 & 4 & 7 & 10 & 12 & 14 \\ \hline \hline
\textbf{p99 Latency (ms)} & 2.4 & 5.1 & 11.9 & 17.1 & 28.7 & 24.2 \\ \hline
\end{tabular}
\caption{p99 per-frame planning latency in milliseconds.}
\label{tab:latency_planning}
\end{table}

\section{Related Work}
\label{sec:related_work}

\parab{Connected Autonomous Vehicles.} Network connectivity in vehicles has opened up large avenues for research. A large body of work~\cite{CVSurvey} has explored wireless technologies and standards (such as DSRC) for vehicle-to-vehicle and vehicle-to-infrastructure communication. Connected autonomous vehicles have also inspired proposals for cooperative perception~\cite{AVR, emp, vieye, vips, vimap},
collaborative map updates~\cite{CarMap},
and cooperative driving~\cite{dariani2019cooperative} in which autonomous vehicles share information with each other to improve safety and utilization. Some have proposed approaches to offload route planning (but not trajectory planning) to the cloud~\cite{guanetti2018control}. Others explore  \textit{platooning}~\cite{schindler2018dynamic} in which vehicles collaboratively and dynamically form platoons to enable smooth traffic flows. Beyond inter-vehicle collaboration, several proposals have explored infrastructure support for connected autonomous vehicles, with infrastructure augmenting perception~\cite{ravipati19:_vision_based_local_infras_enabl_auton,gopalswamy18:_infras_enabl_auton}, or delivering traffic light status~\cite{schindler2018dynamic}. Other work focuses on infrastructure-assisted traffic management at intersections~\cite{AIMSurvey}.
\sysname goes beyond this body of work by demonstrating the feasibility of decoupling both perception and planning from vehicular control.

\parab{Infrastructure LiDAR-based Perception.}  Prior work has explored using infrastructure LiDAR to detect pedestrians~\cite{ped_veh_detection} and road features such as lanes and drivable surfaces~\cite{background_filtering,lanedetection}, and to warn vehicles of impending collisions~\cite{Aycard2011}. One work~\cite{lidar_point_registration_remote_sensing} proposes a genetic algorithm based LiDAR alignment, but unlike \sysname, it has not explored the efficacy of an entire perception pipeline built on top of LiDAR fusion.

\parab{Point Cloud Alignment.} \sysname's alignment builds upon point cloud registration techniques~\cite{point_to_plane}. Prior work has matched features~\cite{persistent_feature_histograms}; these don't work well for \sysname, where LiDARs capture the scene from very different perspectives.

\parab{Deep Neural Nets for 3D Detection and Tracking.}
For vehicle-mounted LiDARs, prior work has developed expensive neural nets for point cloud based detection~\cite{VoxelNet, Yang_2019_ICCV} and tracking~\cite{Qi_2020_CVPR,fast_and_furious}.
These are for vehicle-mounted LiDAR and are computationally expensive; 
\sysname exploits static LiDARs and can use more efficient algorithms, \secref{sec:eval-microbenchmark}.

\parab{Motion Estimation.} Heading and speed can be estimated using DNNs~\cite{fast_and_furious}, SLAM~\cite{ORB-SLAM2}, or visual odometry~\cite{rosinol2020kimera}. \sysname uses a lightweight technique since it relies on static LiDARs.

\section{Conclusions}
\label{s:conclusions}

Fast cooperative infrastructure perception using multiple infrastructure sensors can enable novel automotive and outdoor mixed reality applications.
\sysname contains a suite of algorithms that generates cooperative perception outputs with a p99 latency of 100~ms, while still being as accurate as the state-of-the-art.
It achieves this using careful alignment, reuse of visual abstractions, and systems optimizations including accelerator offload.
When used to augment vehicle perception, it can improve safety.
When used in conjunction with offloaded perception, it can increase traffic throughput at intersections.
\bibliography{references.bib}

\begin{thebibliography}{10}
\providecommand{\url}[1]{#1}
\csname url@samestyle\endcsname
\providecommand{\newblock}{\relax}
\providecommand{\bibinfo}[2]{#2}
\providecommand{\BIBentrySTDinterwordspacing}{\spaceskip=0pt\relax}
\providecommand{\BIBentryALTinterwordstretchfactor}{4}
\providecommand{\BIBentryALTinterwordspacing}{\spaceskip=\fontdimen2\font plus
\BIBentryALTinterwordstretchfactor\fontdimen3\font minus \fontdimen4\font\relax}
\providecommand{\BIBforeignlanguage}[2]{{%
\expandafter\ifx\csname l@#1\endcsname\relax
\typeout{** WARNING: IEEEtran.bst: No hyphenation pattern has been}%
\typeout{** loaded for the language `#1'. Using the pattern for}%
\typeout{** the default language instead.}%
\else
\language=\csname l@#1\endcsname
\fi
#2}}
\providecommand{\BIBdecl}{\relax}
\BIBdecl

\bibitem{BaiduSpecs}
Baidu, ``Apollo: Open source autonomous driving,'' 2017.

\bibitem{Autoware}
K.~Miura, S.~Tokunaga, N.~Ota \emph{et~al.}, ``Autoware toolbox: Matlab/simulink benchmark suite for ros-based self-driving software platform,'' in \emph{RSP}, 2019.

\bibitem{autocast}
H.~Qiu, P.~Huang, N.~Asavisanu \emph{et~al.}, ``Autocast: Scalable infrastructure-less cooperative perception for distributed collaborative driving,'' in \emph{MobiSys}, 2022.

\bibitem{vips}
S.~Shi, J.~Cui, Z.~Jiang \emph{et~al.}, ``Vips: Real-time perception fusion for infrastructure-assisted autonomous driving,'' in \emph{MobiCom}, 2022.

\bibitem{ouster_chattanooga}
``{H}ow {C}hattanooga is {A}chieving {V}ision {Z}ero with {O}uster {L}idar,'' \url{https://ouster.com/blog/how-chattanooga-is-achieving-vision-zero-with-ouster-lidar/}.

\bibitem{seoulrobotics}
``Inside seoul robotics's contrarian approach to autonomous vehicle tech,'' \url{https://techcrunch.com/2022/09/22/seoul-robotics-aims-to-automate-vehicles-movement-via-its-3d-sensor-platform-closes-25m-funding/}.

\bibitem{sslidars}
D.~Trends, ``{A} {S}elf-{D}riving {C}ar in {E}very {D}riveway? {S}olid-{S}tate {L}idar is the {K}ey,'' \url{https://www.digitaltrends.com/cars/solid-state-lidar-for-self-driving-cars/}, 2018.

\bibitem{GDCE}
``Google distributed cloud edge,'' \url{https://cloud.google.com/distributed-cloud-edge}, 2022.

\bibitem{Qualcomm5g}
``{Q}ualcomm 5{G},'' \url{https://www.qualcomm.com/invention/5g}, 2020.

\bibitem{lin18:_archit_implic_auton_drivin}
S.~Lin, Y.~Zhang, C.~Hsu \emph{et~al.}, ``The architectural implications of autonomous driving: Constraints and acceleration,'' in \emph{ASPLOS}, 2018.

\bibitem{geiger2013vision}
A.~Geiger, P.~Lenz, C.~Stiller \emph{et~al.}, ``Vision meets robotics: The kitti dataset,'' \emph{IJRR}, 2013.

\bibitem{google_maps}
``{G}oogle {M}aps,'' \url{www.google.com/maps}.

\bibitem{chen91:_objec}
Y.~Chen and G.~G. Medioni, ``Object modeling by registration of multiple range images,'' in \emph{ICRA}, 1991.

\bibitem{rusu09:_fast_point_featur_histog_fpfh}
R.~B. Rusu, N.~Blodow, and M.~Beetz, ``Fast point feature histograms {(FPFH)} for 3d registration,'' in \emph{ICRA}, 2009.

\bibitem{duchon2012some}
F.~Duchon, M.~Dekan, L.~Jurisica, and A.~Vitko, ``Some applications of laser rangefinder in mobile robotics,'' \emph{Journal of Control Engineering and applied informatics}, vol.~14, no.~2, pp. 50--57, 2012.

\bibitem{derpanis2010overview}
K.~G. Derpanis, ``Overview of the ransac algorithm,'' \emph{Image Rochester NY}, 2010.

\bibitem{vaquero2017deconvolutional}
V.~Vaquero, I.~del Pino, F.~Moreno-Noguer \emph{et~al.}, ``Deconvolutional networks for point-cloud vehicle detection and tracking in driving scenarios,'' in \emph{ECMR}, 2017.

\bibitem{garcia2008fast}
V.~Garcia, E.~Debreuve, and M.~Barlaud, ``Fast k nearest neighbor search using gpu,'' in \emph{CVPR Workshops}, 2008.

\bibitem{umeyama}
S.~Umeyama, ``Least-squares estimation of transformation parameters between two point patterns,'' \emph{IEEE Comput. Archit. Lett.}, 1991.

\bibitem{gao2018gpu}
M.~Gao, X.~Wang, K.~Wu \emph{et~al.}, ``Gpu optimization of material point methods,'' \emph{TOG}, 2018.

\bibitem{kashani2015review}
A.~G. Kashani, M.~J. Olsen, C.~E. Parrish \emph{et~al.}, ``A review of lidar radiometric processing: From ad hoc intensity correction to rigorous radiometric calibration,'' \emph{Sensors}, 2015.

\bibitem{Ouster}
Ouster, ``{O}uster {L}i{DAR},'' \url{https://ouster.com/}, 2020.

\bibitem{ester1996density}
M.~Ester, H.-P. Kriegel, J.~Sander \emph{et~al.}, ``A density-based algorithm for discovering clusters in large spatial databases with noise.'' in \emph{KDD}, 1996.

\bibitem{pclbbox}
``Find minimum oriented bounding box of point cloud,'' \url{http://codextechnicanum.blogspot.com/2015/04/find-minimum-oriented-bounding-box-of.html}, 2015.

\bibitem{NHTSA-precrash}
W.~G. Najm, R.~Ranganathan, G.~Srinivasan \emph{et~al.}, ``Description of light-vehicle pre-crash scenarios for safety applications based on vehicle-to-vehicle communications,'' US. NHTSA, Tech. Rep., 2013.

\bibitem{carlachallenge}
CarLA, ``Carla autonomous driving challenge,'' \url{https://carlachallenge.org/}.

\bibitem{vrf}
K.~Nawaz~Khan, A.~Khalid, T.~Yash, K.~Dantu, and F.~Ahmad, ``V{R}{F}: {V}ehicle {R}oad-side {P}oint {C}loud {F}usion,'' in \emph{MobiSys}, 2024.

\bibitem{paden2016survey}
B.~Paden, M.~Cap, S.~Z. Yong \emph{et~al.}, ``A survey of motion planning and control techniques for self-driving urban vehicles,'' \emph{IEEE {T-IV}}, 2016.

\bibitem{traffic-light-free}
R.~Tachet, P.~Santi, S.~Sobolevsky \emph{et~al.}, ``Revisiting street intersections using slot-based systems,'' \emph{PLOS ONE}, 2016.

\bibitem{SIPP}
M.~Phillips and M.~Likhachev, ``Sipp: Safe interval path planning for dynamic environments,'' in \emph{ICRA}, 2011.

\bibitem{quigley2009ros}
M.~Quigley, K.~Conley, B.~Gerkey \emph{et~al.}, ``Ros: an open-source robot operating system,'' in \emph{ICRA workshop on OSS}, 2009.

\bibitem{sipp_impl}
Whoenig, ``{L}ibrary with {S}earch {A}lgorithms for {T}ask and {P}ath {P}lanning for {M}ulti {R}obot/{A}gent {S}ystems,'' \url{https://github.com/whoenig/libMultiRobotPlanning}.

\bibitem{busch2022lumpi}
S.~Busch, C.~Koetsier, J.~Axmann \emph{et~al.}, ``Lumpi: The leibniz university multi-perspective intersection dataset,'' in \emph{IV}.\hskip 1em plus 0.5em minus 0.4em\relax IEEE, 2022.

\bibitem{nhsta2013}
W.~G. Najm, R.~Ranganathan, G.~Srinivasan \emph{et~al.}, ``Description of light-vehicle pre-crash scenarios for safety applications based on vehicle-to-vehicle communications,'' US. NHTSA, Tech. Rep., 2013.

\bibitem{mot_metric}
K.~Bernardin and R.~Stiefelhagen, ``Evaluating multiple object tracking performance: the clear mot metrics,'' \emph{EURASIP}, 2008.

\bibitem{vloam}
J.~Zhang and S.~Singh, ``Visual-lidar odometry and mapping: Low-drift, robust, and fast,'' in \emph{ICRA}, 2015.

\bibitem{mono3dt}
H.-N. Hu, Q.-Z. Cai, D.~Wang \emph{et~al.}, ``Joint monocular 3d vehicle detection and tracking,'' in \emph{ICCV}, 2019.

\bibitem{cvivsic2022soft2}
I.~Cvi{\v{s}}i{\'c}, I.~Markovi{\'c}, and I.~Petrovi{\'c}, ``Soft2: Stereo visual odometry for road vehicles based on a point-to-epipolar-line metric,'' \emph{IEEE Tran. on Robotics}, 2022.

\bibitem{intentNet}
S.~Casas, W.~Luo, and R.~Urtasun, ``{IntentNet: Learning to Predict Intention from Raw Sensor Data},'' in \emph{CoRL}.\hskip 1em plus 0.5em minus 0.4em\relax PMLR, 2018.

\bibitem{car_length}
``Average car length,'' \url{https://anewwayforward.org/average-car-length/}.

\bibitem{lane_width}
``Average lane width,'' \url{https://safety.fhwa.dot.gov/geometric/pubs/mitigationstrategies/chapter3/3_lanewidth.cfm}.

\bibitem{three_second}
``3-second rule for safe following distance,'' \url{https://www.travelers.com/resources/auto/travel/3-second-rule-for-safe-following-distance}.

\bibitem{mobileye}
S.~Shalev-Shwartz, S.~Shammah, and A.~Shashua, ``Safe, multi-agent, reinforcement learning for autonomous driving,'' \emph{arXiv}, 2016.

\bibitem{AIMSurvey}
M.~Khayatian, M.~Mehrabian, E.~Andert \emph{et~al.}, ``A survey on intersection management of connected autonomous vehicles,'' \emph{ACM Trans. Cyber-Phys. Syst.}, 2020.

\bibitem{intelligent_tl}
F.~Ahmad, S.~A. Mahmud, and F.~Z. Yousaf, ``Shortest processing time scheduling to reduce traffic congestion in dense urban areas,'' \emph{IEEE Trans. Syst. Man Cybern. Syst.}, 2016.

\bibitem{natco}
``Signal cycle lengths,'' \url{https://nacto.org/publication/urban-street-design-guide/intersection-design-elements/traffic-signals/signal-cycle-lengths/}.

\bibitem{CVSurvey}
J.~E. {Siegel}, D.~C. {Erb}, and S.~E. {Sarma}, ``A survey of the connected vehicle landscape—architectures, enabling technologies, applications, and development areas,'' \emph{IEEE T-ITS}, 2018.

\bibitem{AVR}
H.~Qiu, F.~Ahmad, F.~Bai \emph{et~al.}, ``Avr: Augmented vehicular reality,'' in \emph{MobiSys}, 2018.

\bibitem{emp}
X.~Zhang, A.~Zhang, J.~Sun \emph{et~al.}, ``Emp: edge-assisted multi-vehicle perception,'' in \emph{MobiCom}, 2021.

\bibitem{vieye}
Y.~He, L.~Ma, Z.~Jiang \emph{et~al.}, ``Vi-eye: Semantic-based 3d point cloud registration for infrastructure-assisted autonomous driving,'' in \emph{MobiCom}.\hskip 1em plus 0.5em minus 0.4em\relax ACM, 2021.

\bibitem{vimap}
Y.~He, C.~Bian, J.~Xia \emph{et~al.}, ``Vi-map: Infrastructure-assisted real-time hd mapping for autonomous driving,'' in \emph{MobiCom}, 2023.

\bibitem{CarMap}
F.~Ahmad, H.~Qiu, R.~Eells \emph{et~al.}, ``Carmap: Fast 3d feature map updates for automobiles,'' in \emph{NSDI}, 2020.

\bibitem{dariani2019cooperative}
R.~Dariani and J.~Schindler, ``Cooperative strategical decision and trajectory planning for automated vehicle in urban areas,'' in \emph{ICVES}, 2019.

\bibitem{guanetti2018control}
J.~Guanetti, Y.~Kim, and F.~Borrelli, ``Control of connected and automated vehicles: State of the art and future challenges,'' \emph{Annual Reviews in Control}, 2018.

\bibitem{schindler2018dynamic}
J.~Schindler, R.~Dariani, M.~Rondinone \emph{et~al.}, ``Dynamic and flexible platooning in urban areas,'' in \emph{AAET}, 2018.

\bibitem{ravipati19:_vision_based_local_infras_enabl_auton}
D.~Ravipati, K.~Chour, A.~Nayak \emph{et~al.}, ``Vision based localization for infrastructure enabled autonomy,'' in \emph{ITSC}, 2019.

\bibitem{gopalswamy18:_infras_enabl_auton}
S.~Gopalswamy and S.~Rathinam, ``Infrastructure enabled autonomy: {A} distributed intelligence architecture for autonomous vehicles,'' in \emph{{IEEE} IV}, 2018.

\bibitem{ped_veh_detection}
J.~Zhao, H.~Xu, H.~Liu \emph{et~al.}, ``{Detection and tracking of pedestrians and vehicles using roadside LiDAR sensors},'' \emph{Transp. Res. Part C Emerg.}, 2019.

\bibitem{background_filtering}
J.~Wu, H.~Xu, and J.~Zheng, ``{Automatic background filtering and lane identification with roadside LiDAR data},'' in \emph{ITSC}, 2017.

\bibitem{lanedetection}
A.~B. Hillel, R.~Lerner, D.~Levi, and G.~Raz, ``Recent progress in road and lane detection: a survey,'' \emph{Mach. Vis. Appl.}, 2014.

\bibitem{Aycard2011}
O.~Aycard, ``{Intersection Safety Using Lidar and Stereo Vision Sensors on a Demonstrator Vehicle},'' \emph{Transportation}, 2011.

\bibitem{lidar_point_registration_remote_sensing}
R.~Yue, H.~Xu, J.~Wu \emph{et~al.}, ``{Data registration with ground points for roadside LiDAR sensors},'' \emph{Remote Sensing}, 2019.

\bibitem{point_to_plane}
Y.~{Chen} and G.~{Medioni}, ``Object modeling by registration of multiple range images,'' in \emph{ICRA}, 1991.

\bibitem{persistent_feature_histograms}
R.~B. {Rusu}, N.~{Blodow}, Z.~C. {Marton}, and M.~{Beetz}, ``Aligning point cloud views using persistent feature histograms,'' in \emph{IROS}, 2008.

\bibitem{VoxelNet}
Y.~Zhou and O.~Tuzel, ``Voxelnet: End-to-end learning for point cloud based 3d object detection,'' in \emph{CVPR}, 2018.

\bibitem{Yang_2019_ICCV}
Z.~Yang, Y.~Sun, S.~Liu, X.~Shen, and J.~Jia, ``Std: Sparse-to-dense 3d object detector for point cloud,'' in \emph{ICCV}, 2019.

\bibitem{Qi_2020_CVPR}
H.~Qi, C.~Feng, Z.~Cao, F.~Zhao, and Y.~Xiao, ``P2b: Point-to-box network for 3d object tracking in point clouds,'' in \emph{CVPR}, 2020.

\bibitem{fast_and_furious}
W.~Luo, B.~Yang, and R.~Urtasun, ``{Fast and Furious: Real Time End-to-End 3D Detection, Tracking and Motion Forecasting With a Single Convolutional Net},'' in \emph{CVPR}, 2018.

\bibitem{ORB-SLAM2}
R.~Mur-Artal and J.~D. Tard{\'o}s, ``Orb-slam2: An open-source slam system for monocular, stereo, and rgb-d cameras,'' \emph{IEEE Trans. Robotics}, 2017.

\bibitem{rosinol2020kimera}
A.~Rosinol, M.~Abate, Y.~Chang \emph{et~al.}, ``Kimera: an open-source library for real-time metric-semantic localization and mapping,'' in \emph{ICRA}, 2020.

\end{thebibliography}

\label{lastpage}

\end{document}